\documentclass[sensors,article,accept,moreauthors,pdftex]{Definitions/mdpi}

\firstpage{1}
\pubvolume{20}
\issuenum{15}
\articlenumber{4097}
\pubyear{2020}
\copyrightyear{2020}
\history{Received: 8 May 2020; Accepted: 21 July 2020; Published: 23 July 2020}
\updates{yes} 

\usepackage[labelformat=simple]{subcaption}

\DeclareCaptionLabelFormat{subcaptionlabel}{\normalfont(\textbf{#2}\normalfont)}
\Title{Fail-Aware LIDAR-Based Odometry for Autonomous Vehicles}

\Author{{Iván García Daza} 
*\href{https://orcid.org/0000-0001-8940-6434}{\orcidicon}, {Mónica Rentero, Carlota Salinas Maldonado\href{https://orcid.org/0000-0003-3211-1752}{\orcidicon}, Ruben Izquierdo Gonzalo, Noelia Hernández Parra, Augusto Ballardini\href{https://orcid.org/0000-0001-6688-5081}{\orcidicon} and David Fernandez Llorca}}

\AuthorNames{Iván García Daza}

\address[1]{%
{Computer Engineering Department, Universidad de Alcalá}
, 28805 Alcalá de Henares, Spain; monica.rentero@edu.uah.es (M.R.); carlota.salinasmaldo@uah.es (C.S.M.); ruben.izquierdo@uah.es (R.I.G.); noelia.hernandez@uah.es (N.H.P.); augusto.ballardini@uah.es (A.B.); david.fernandezl@uah.es (D.F.L.)}

\corres{\hangafter=1 \hangindent=1.05em \hspace{-0.82em}Correspondence: ivan.garciad@uah.es; Tel.: +34-918-856-622}

\abstract{Autonomous driving systems are set to become a reality in transport systems and, so, maximum acceptance is being sought among users. Currently, the most advanced architectures require driver intervention when functional system failures or critical sensor operations take place, presenting problems related to driver state, distractions, fatigue, and other factors that prevent safe control. Therefore, this work presents a redundant, accurate, robust, and scalable LiDAR odometry system with fail-aware system features that can allow other systems to perform a safe stop manoeuvre without driver mediation. All odometry systems have drift error, making it difficult to use them for localisation tasks over extended periods. For this reason, the paper presents an accurate LiDAR odometry system with a fail-aware indicator. This indicator estimates a time window in which the system manages the localisation tasks appropriately. The odometry error is minimised by applying a dynamic 6-DoF model and fusing measures based on the Iterative Closest Points (ICP), environment feature extraction, and Singular Value Decomposition (SVD) methods. The obtained results are promising for two reasons: First, in the KITTI odometry data set, the ranking achieved by the proposed method is twelfth, considering only LiDAR-based methods, where its translation and rotation errors are $1.00\%$ and 0.0041 {deg}
/m, respectively. Second, the encouraging results of the fail-aware indicator demonstrate the safety of the proposed LiDAR odometry system. The results depict that, in order to achieve an accurate odometry system, complex models and measurement fusion techniques must be used to improve its behaviour. Furthermore, if an odometry system is to be used for redundant localisation features, it must integrate a fail-aware indicator for use in a safe~manner.}


\begin{document}



\section{Introduction}
\label{section:introduction}
\vspace{-6pt}

\subsection{Motivation}

At present, the~concept of autonomous driving is becoming more and more popular. Therefore, new techniques are being developed and researched to help consolidate the reality of implementing the concept. As~systems become autonomous, their safety must be improved to increase user acceptance. Consequently, it is necessary to integrate intelligent fault detection systems that guarantee the security of passengers and people in the environment. Sensors, perception, localisation, or~control systems are essential elements for their development. However, they are also susceptible to failures and it is necessary to have fail-x systems, which prevent undesired or fatal actions. A~fail-x system combines the following features: redundancy in design (fail-operational), ability to plan emergency manuevers and undertake safe stops (fail-safe), and~monitoring the status of their sensors to detect failures or malfunctions in them (fail-aware). At~present, in~an urban environment where there are increasingly complex traffic elements such as multiple intersections, complex lane roundabouts, or~tunnels, a~localisation system based only on GPS may pose problems. Thus, autonomous driving will be a closer reality when LiDAR or Visual odometry systems are integrated to cover fail-operational functions. However, fail-aware behaviour has to be integrated into the global system~also.

At present, the~Global Positioning System (GPS) performs the main tasks of localisation due to its robustness and accuracy. However,  GPS coverage problems derived from structural elements of the road (tunnels), GPS multi-path in urban areas, or~failure in its operation, mean that this technology does not meet  the necessary localisation requirements in 100\% of use-cases, which~makes it essential to design redundant systems based on LiDAR odometry~\cite{Intelligent_Feature_Selection}, Visual odometry~\cite{parra_sotelo_llorca_ocana_2010}, Inertial~Navigation Systems (INS) \cite{milanes_naranjo_gonzalez_alonso_de_pedro_2008}, Wifi~\cite{8917290}, or~a combination of the above, including digital maps~\cite{6192327}. However, LiDAR~and Visual odometry systems suffer from a non-constant temporal drift, where the characteristics of the environment and the algorithm behaviour are determinants that improve or worsen this drift. Therefore, it is necessary to introduce, for~those systems that have a non-constant temporal drift, a~fail-aware indicator to discern when these can be~used.

\subsection{Problem~Statement}

Safe behaviour in a vehicle's control and navigation systems depends mostly on the redundancy and failure detections that these present. At~the moment, when GPS-based localisation fails, either~temporarily or permanently, the~LiDAR and Visual odometry systems can start as redundant localisation systems, mitigating the erroneous behaviour of the GPS localisation. Redundant~localisation based on 3D mapping techniques can be applied, as~well. However, this~is currently more widespread in robotic applications, as~the 3D map accuracy in open environments is decisive for localisation tasks. However, companies such as Mapillary and Here have presented promising results for 3D map accuracy. Why is it challenging to build an accurate 3D map when relying only on GPS localisation? It is because the GPS angular error feature of market devices is close to $10^{-3}$ rad. This~feature can place a 3D object with an error of 0.01 m when the object distance from the sensor is 100 m.

So, in~the case of integrating redundancy into the localisation system with an odometry alternative, a~fail-aware indicator has to be integrated into the odometry system, as~a consequence of the non-constant drift error, in~order for it to be used as a redundant system. The~fail-aware indicator could be based on an estimated time window that satisfies a localisation error below the minimum requirements to planned emergency manoeuvring and placing the vehicle in a safe spot. Several~alternatives can be presented to implement the fail-aware indicator. The~first is to set a fixed time window in which the system is used. The~second alternative is an adaptive time window, which~is evaluated dynamically in the continuous localisation process to find the maximum time in which the redundant system can be used. At~present, there have been no recent works focused on fail-aware LIDAR-based odometry for autonomous~vehicles.

Therefore, it is necessary to look for an odometry process that maximises the time in exceeding the threshold that leads the system to a failure state and, for~that purpose, we propose a robust, scalable, and~precise localisation design that minimises the error in each iteration. Multiple measurement fusion techniques from both global positioning systems and odometry systems are used to make the system robust. Bayesian filtering guarantees an optimal fusion between the observation techniques applied in the odometry systems and improves the localisation accuracy by integrating (mostly kinematic) models of the vehicle's displacement, having either three or six degrees of freedom (DoF). The~LiDAR odometry is based exclusively on the observations of the LiDAR sensor, where the emission of near-infrared pulses and the measurement of the reflection time allows us to represent the scene with a set of 3D points, called a Point Cloud. Thus, given a temporal sequence of measurements, we obtain the homogeneous transformation, rotation, and~translation corresponding to two consecutive time instants, by~applying iterative registering and optimisation methods. However, this process alone provides incorrect homogeneous transformations if the scene presents moving objects and, so, solutions based on feature detection  must be explored in order to mitigate possible~errors.

\subsection{Contributions}

\begin{figure}[H]
\centering
\includegraphics[width=0.9\linewidth]{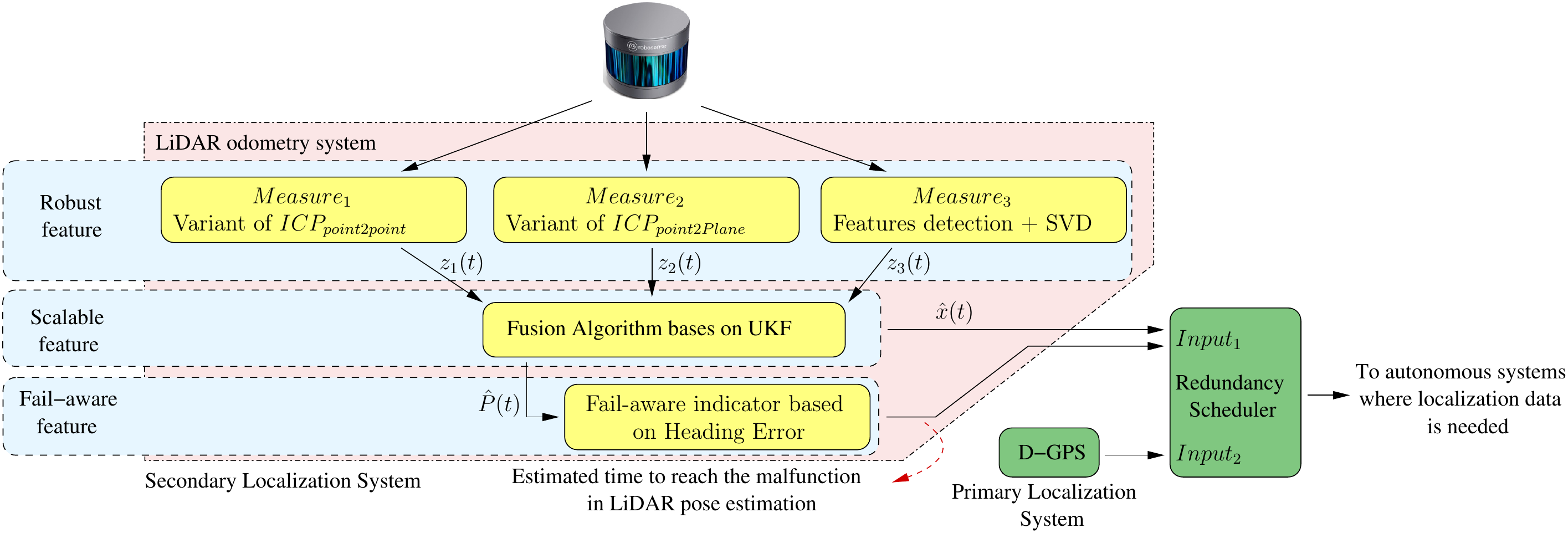}
\caption{General diagram. The~developed blocks are represented in yellow. The~horizontal blue strips represent the main features of the odometry system. A~framework where the LiDAR odometry system can be integrated within the autonomous driving cars topic is depicted with green blocks, such as a secondary localisation~system.}
\label{fig:generalDiagram}
\end{figure}

The factors described previously motivated us to develop an accurate LiDAR odometry system with a fail-aware indicator ensuring its proper use as a redundant localisation system for autonomous vehicles, as~shown in Figure~\ref{fig:generalDiagram}. The~accurate LiDAR odometry architecture is based on robust and scalable features. The~architecture has a robust measurement topology as it integrates three measurement algorithms, two of which are based on Iterative Closest Point (ICP) variants, and~the third one is based on non-mobile scene feature extraction and Singular Value Decomposition (SVD). Furthermore, our work proposes a scalable architecture to integrate a fusing block, which relies on the UKF scheme. Another factor taken into account to enhance the odometry accuracy was to incorporate a 6-DoF motion model based on vehicle dynamics and kinematics within the filter, where the variables of pitch and roll play a crucial impact on the precision. The~proposed scalable architecture allows us to fuse any position measurement system or integrate into the LiDAR odometry system new measurement algorithms in a natural way. A~fail-aware indicator based on the vehicle heading error is another contribution to the state-of-the-art. The~fail-aware indicator introduces, in~the system output, an~estimated time to reach system malfunction, which enables other systems to take it into~consideration.

The global system is validated by processing KITTI odometry data set sequences and evaluating the error committed in each of the available sequences, allowing for comparison with other state-of-the-art techniques. The~variability in the available scenes allows us to validate the fail-aware functionality, by~comparing sequences with low operating error with those with higher error, and~observing how the temporal estimation factor increases for those sequences with worse~results.

The rest of the paper is comprised of the following sections: Section~\ref{sec:related_works} presents the state-of-the-art in the areas of LiDAR odometry and vehicle motion modelling. Section~\ref{sec:model} details the integrated 6-DoF model, while Section~\ref{sec:filteringUKF} explains the global software architecture of the work, as~well as the methodology applied to fuse the evaluated measures. Then, Section~\ref{sec:measurements} describes the details of the proposed measurement systems. Section~\ref{sec:failAware} describes the methodology to make the systems fail-aware. Section~\ref{section:experimentalAnalysis} describes, lists, and~compares the results obtained by the developed system. Finally, Section~\ref{sec:conclusions} presents our conclusions and proposes a description of future work in the fields of odometry techniques, 3D mapping, and~fail-aware~systems.

\section{Related~Works}
\label{sec:related_works}

Many contemporary conceptual systems for autonomous driving require precise localisation systems. The~geo-referenced localisation system does not usually satisfy such precision,  as~there are scenarios (e.g., tunnel or forest scenarios) where the localisation provided by GPS is not correct or has low accuracy, leading to safety issues and non-robust behaviours. For~these reasons, GPS-based localisation techniques do not satisfy all the use-cases of driverless vehicles, and~it is therefore mandatory to integrate other technologies into the final localisation system. Visual odometry systems could be candidates for such a technology, but~scenarios with few characteristics or extreme environmental conditions lead to their non-robust performance, although~considerable progress has been made in this area, as~described in~\cite{Scaramuzza,Pavan}. Nevertheless, LiDAR odometry systems mitigate part of the visual-related problems, but~real-time features or accuracy in the algorithm remain issues. In~the same way as Visual Odometry, significant advances and results have been obtained, in~the last few years, in~this~topic.

The general challenge in odometry is to evaluate a vehicle's movement at all times without error, in~order to obtain zero global localisation error; however, this issue is not reachable as the odometry measurement commits a small error in each iteration.  These systems therefore have the weakness of drift error over time due to the accumulation of iterative errors; which is a typical integral problem. Drift error is a function of path length or elapsed time. However, these techniques have advanced in the last few decades, due to the improvement of sensor accuracies, achieving small errors (as presented in~\cite{vloam}).

LIDAR odometry is based on the procedures of point registration and optimisation between two consecutive frames. Many works have been inspired by these techniques, but~they have the disadvantage of not ensuring a global solution, introducing errors in their performances. These~techniques are called Iterative Closest Points (ICP), many of which have been described in~\cite{Rusinkiewicz}, where modifications affecting all phases of the ICP algorithm  were analysed, from~the selection and registration of points to the minimisation approach. The~most widely used are ICP point-to-point {$\text{ICP}_{\text{p2p}}$} \cite{Xiaojing, LOAM} and ICP point-to-plane {$\text{ICP}_{\text{p2p}}$} \cite{icpp2P, Sun}. For~example,~presented a point-to-point ICP algorithm based on two stages \cite{Xiaojing}, in~order to improve its exactness. Initially, a~coarse registration stage based on KD-tree is applied to solve the alignment problem. Once the first transformation is applied, a~second fine-recording stage based on KD-tree ICP is carried out, in~order to solve the convergence problem more~accurately.

Several optimisation techniques have been proposed for use when the cost function is established. Obtaining a rigid transformation is one of the most commonly used schemes, as~has been detailed in
the simplest ICP case~\cite{Paul}, as~well as in more advanced variants such as CPD~\cite{Myronenko}. This is easily achieved by decoupling the rotation and translation, obtaining the first using algebraic tools such as SVD (Singular Value Decomposition), while the second term is simply the mean/average translation. Other proposals, such as LM-ICP~\cite{Andrew} or~\cite{LOAM}, perform a Levenberg--Marquardt approach to add robustness to the process. Finally, optimisation techniques such as gradient descent have been used in distribution-to-distribution methods like D2D NDT~\cite{Todor}.

In order to increase robustness and computational performance, interest point descriptors for point clouds have  recently been proposed. General point cloud or 2D depth maps are two general approaches to achieve this. The~latter may include curvelet features, as~analysed in~\cite{Ahuja}, assuming the range data is dense and a single viewpoint is used in order to capture the point cloud. However, it~may not perform accurately for a moving LiDAR---the objective of this paper. In~a general point cloud approach, Fast Point Feature Histograms (FPFH) \cite{Rusu} and Integral Volume Descriptors (IVD)~\cite{Zhou} are two feature-based global registration proposals of interest. The~first one generates feature histograms, which have demonstrated great results in vision object detection problems, using such techniques as Histogram of Oriented Gradients (HOG), by~means of computing some statistics about a point's neighbours relative positions and estimated surface normals. Feature histograms have shown the best IVD performances and surface curvature estimates. However, neither of these methods offer reliable results in sparse point clouds and are slow to~compute.

Once one correspondence has been established, using features instead of proximity, it can be used to initialize ICP techniques in order to improve their results. As~described in previous paragraphs, this transformation can also be found by other techniques, such as PCA or SVD, which are both deterministic procedures. In~order to obtain a transformation, three point correspondences are enough, as~is shown in the proposal we introduce in this document. However, as~many outlier points are typically present in a point cloud (such as those of vegetation), a~random sample consensus (RANSAC) approach is usually used~\cite{Rusu}. Other approaches include techniques tailored to the specific problem, such as the detection of structural elements of the scene~\cite{Chen_Xieyuanli}.

In the field of observations or measurements, there are a large number of methods for measuring the homogeneous transformation between two moments or two point clouds. For~this reason, many~filtering and fusion systems have been applied to improve the robustness of systems. The~two~most widespread techniques to filter measurements are recursive filtering and batch optimisation~\cite{Stefan}. Recursive filtering updates the status probabilistically, using only the latest sensor observations for status prediction and process updates. The~Kalman filter and its variants mostly represent recursive filtering techniques. However, filtering based on batch optimisation maintains a history of observations to evaluate, on~the basis of previous states, the~most probable estimate of the current instant. Both techniques may integrate kinematic and dynamic models of the system under analysis to improve the process of estimating observations. In~the field of autonomous driving, the~authors of~\cite{7995816} justified the importance of applying models in the solution of driving problems, raising the need to work with complex models that correctly filter and fuse~observations.

The best odometry system described in the state-of-the-art is VLOAM~\cite{vloam}, which is based on Visual and LiDAR odometry. It is characterised by being a particularly robust method in the face of an aggressive movement and the occasional lack of visual features. The~method starts with a stage of visual odometry, in~order to obtain a first approximation of the movement. The~final stage is executed with LiDAR odometry. The~results shown applied to a set of ad-hoc tests and the KITTI odometry data set. The~work presented as LIMO~\cite{limo} also aimed to evaluate the movement of a vehicle accurately. Stereo images with LiDAR were  used to provide depth information to the features detected by the cameras. The~process includes a semantic segmentation, which is used to reject and weight characteristic points used for odometry. The~results given were related to the KITTI data set.
On the other hand, the~authors of~\cite{caelo} presented a LiDAR odometry technique that models the projection of the point cloud in a 2D ring structure. Over~the 2D structure, an~unsupervised Convolutional Auto-Encoder (CAE-LO) system detects points of interest in the spherical ring (CAE-2D). It later extracts characteristics from the multi-resolution voxel model using 3D CAE. It was characterised as finding 50\% more points of interest in the point cloud, improving the success rate in the cloud comparison process. To~conclude, the~system described in~\cite{mc2} proposed a real-time laser odometry approach, which presented small drift. The~LiDAR system uses inertial data from the vehicle. The~point clouds are captured in motion, but~are compensated with a novel sweep correction algorithm based on two consecutive laser scans and a local~map.

To the best of our knowledge, there have been no recent works focused on fail-aware LiDAR-based odometry for autonomous~vehicles.

\section{Kinematic and Dynamic Vehicle~Model}
\label{sec:model}

Filters usually leverage mathematical models to better approximate state transitions. In~the field of vehicle modelling, there are two ways to study the movement of a vehicle: with kinematic or dynamic models.
In the field of kinematic vehicle modelling, one of the most-used models is the bicycle model, due to its ease of understanding and simplicity. This model requires knowledge of the slide angle ($\beta$) as well as the front wheel angle ($\delta$) parameters. These variables are usually measured by dedicated car~systems.

\begin{figure}[H]
\centering
\includegraphics[width=0.8\linewidth]{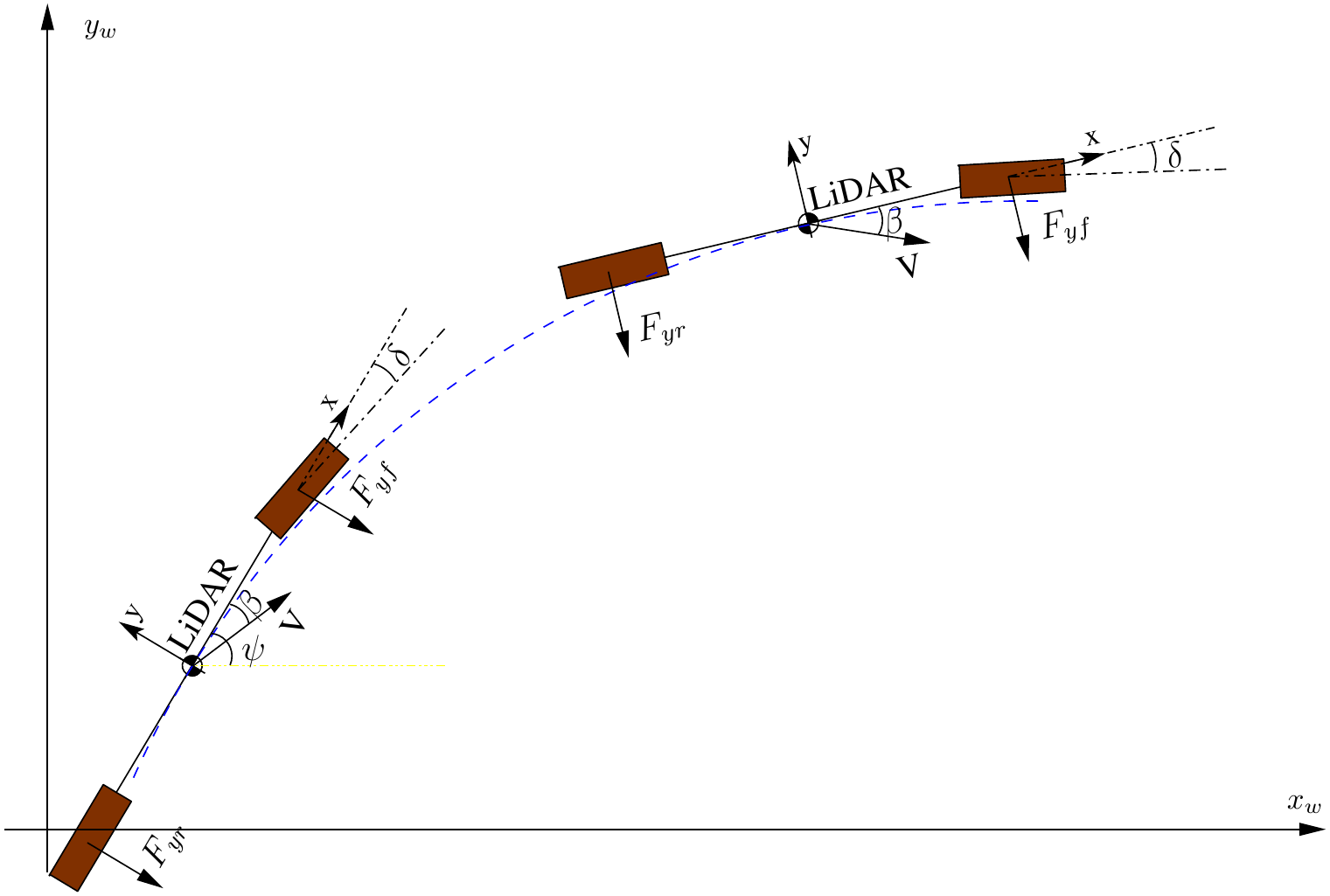}
\caption{Vehicle representation by bicycle model. Using the vehicle reference system, our LiDAR-based odometry process assesses the vehicle forces between two instants of time, allowing for estimation of the $\beta$ and $\delta$ variables. }
\label{fig:bicycle_model}
\end{figure}

In this work, the~variables $\beta$ and $\delta$ are not registered in the data set, so the paper proposes an approach based on a dynamic model to evaluate them. The~method proposed can be used as a redundancy system, replacing dedicated car systems. The~technique relies on the application of LiDAR odometry and the application of vehicle dynamics models where linear and angular forces are taken into account and the variables $\beta$ and $\delta$ are assessed during the car's movement. Figure~\ref{fig:bicycle_model} depicts the actuated forces in the x and y car axes, as~well as the slip angle and the front-wheel angle. Given these variables, the~bicycle model is applied to predict the car's~movement.

From a technical perspective, the~variables $\beta$ and $\delta$ are evaluated using Equations~(\ref{eq:transversalDynamicModel}) and (\ref{eq:SlipAngle}). Equation~(\ref{eq:transversalDynamicModel}) represents Newton's second law, applied on the car's transverse axis in a linear form and on the car's z-axis in an angular form.
\begin{equation}
\begin{matrix}
Translational: && m(\ddot{y}+\dot{\psi}v_x) = F_{yf} + F_{yr} \\
Angular: && I_{z} \ddot{\psi} = l_f F_{yf} - l_r F_{yr}
\end{matrix} ,
\label{eq:transversalDynamicModel}
\end{equation}
where $m$ is the mass of the vehicle, $v_{x}$ is the projection of the speed car vector $V$ on its longitudinal axis $x$, $F_{yf}$ and $F_{yr}$ are the lateral forces produced on the front and rear wheel, $I_z$ is the inertia moment of the vehicle concerning to the z-axis, and~$l_f$ and $l_r$ are the distances from the centre of masses of the front and rear wheels, respectively.

The lateral forces $F_{yf}$ and $F_{yr}$ are, in~turn, functions of characteristic tyre parameters, cornering stiffnesses $C_{\alpha f}$ and $C_{\alpha r}$, the~vehicle chassis configuration $l_f$ and $l_r$, the~linear and angular travel speed to which the vehicle is subjected to $v_x, \dot{\psi}$, the~slip angle $\beta$, and~the turning angle of the front wheel $\delta$, as~shown in Equation~(\ref{eq:SlipAngle}):
\begin{equation}
\begin{matrix}
&& F_{yf} = C_{\alpha f} (\delta - \beta - \frac{l_f \dot{\psi}}{v_x}) \\
&& F_{yr} = C_{\alpha r} (-\beta + \frac{l_r \dot{\psi}}{v_x})
\end{matrix}
\label{eq:SlipAngle}
\end{equation}

\begin{figure}[H]
\centering
\includegraphics[width=0.8\linewidth]{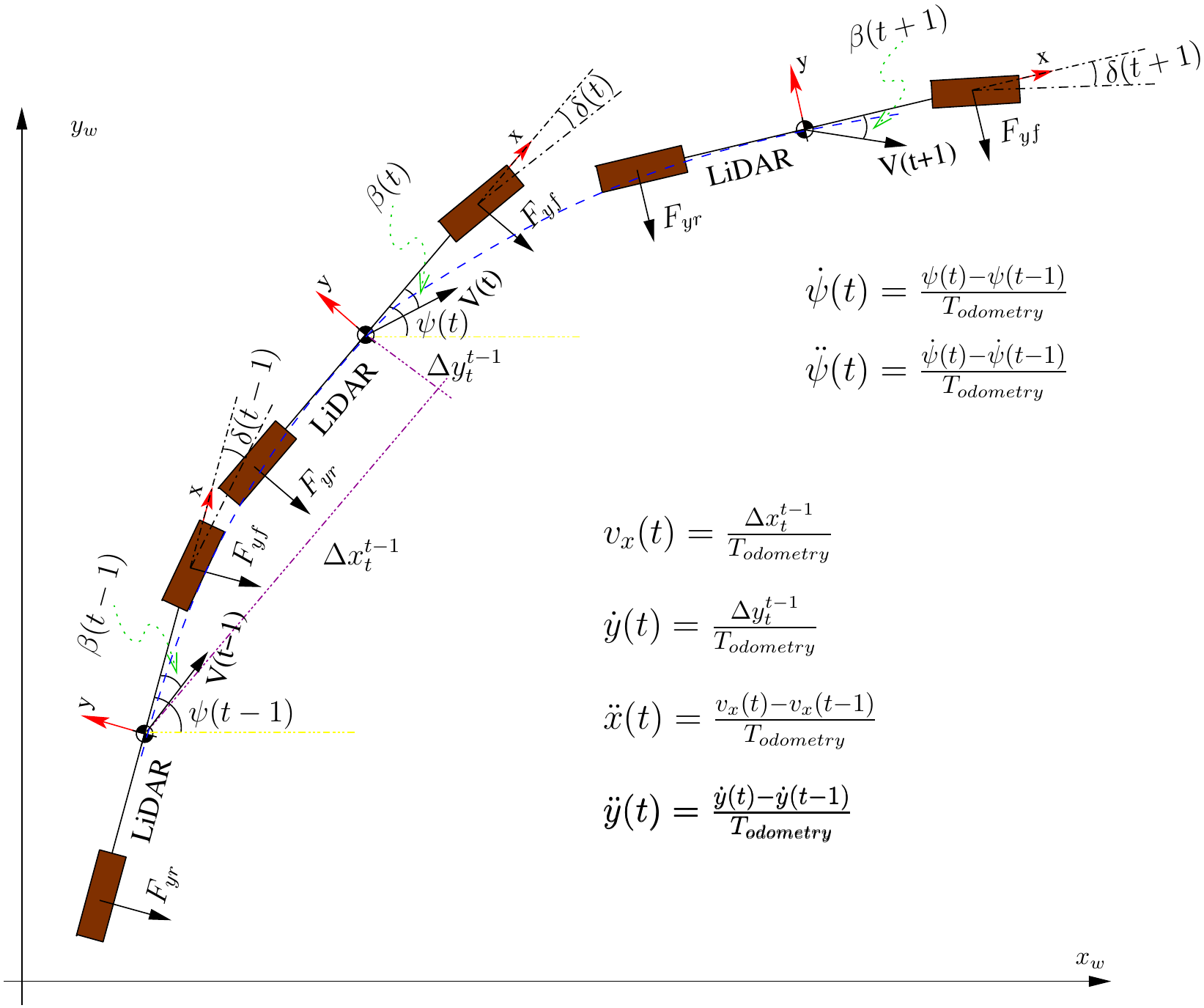}
\caption{Evaluation of variables $\ddot{y}, v_x, \ddot{\psi}, \dot{\psi}$ with LiDAR~odometry.}
\label{fig:odometryProcess}
\end{figure}

Therefore, knowing the above vehicle parameters and assessing the variables $\ddot{y}, v_x, \ddot{\psi}$, and~$\dot{\psi}$ from the LiDAR odometry displacement, with~the method proposed in this work (see Figure~\ref{fig:odometryProcess}), the~variables $\beta$ and $\delta$ can be derived by solving the two-equation system shown in Equation~(\ref{eq:transversalDynamicModel}).

Finally, the~variables $\beta$ and $\delta$ are used in the kinematic bicycle model defined by Equation~(\ref{eq:kinematicModelEquations}) to obtain the speeds $[\dot{X}, \dot{Y}, \dot{\psi}]$ which, in~turn, are used to output the predicted vehicle pose at time $(t+1)$.
\begin{equation}
\begin{matrix}
\dot{X} = V\cos(\psi + \beta) \\
\dot{Y} = V\sin(\psi + \beta) \\
\dot{\psi} = \frac{V\cos(\beta)}{l_f + l_r}(\tan(\delta_f) - \tan(\delta_r))
\end{matrix} .
\label{eq:kinematicModelEquations}
\end{equation}

\begin{figure}[H]
\centering
\includegraphics[width=0.6\linewidth]{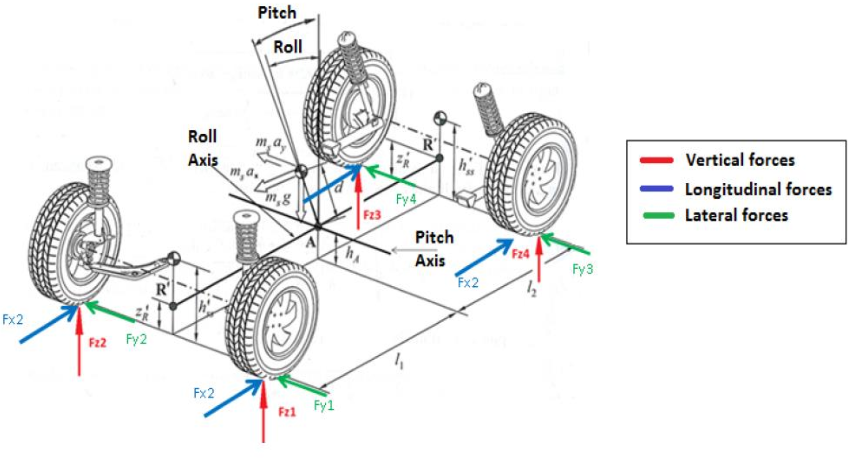}
\caption{Detail of forces and moments applied to the vehicle. The~distance $d$ represented is broken down into $d_{picth}$ and $d_{roll}$, concerning the pitch and roll axes of rotation, respectively. The~figure references~\cite{Obialero2013ARV}.}
\label{fig:rollPitchForces}
\end{figure}

However, the~model mentioned above evaluates the vehicle's motion only in 3-DoF, while~the LiDAR odometry gives us full 6-DoF displacement. Therefore, to~assess the remaining 3-DoF, we~propose to use another dynamic model based on the behaviour of the shock absorbers and the position of the vehicle's mechanical pitch ($\theta$) and roll ($\alpha$) axes; see Figure~\ref{fig:rollPitchForces}. Appling this second dynamical model, we can predict the car's movement in terms of its~6-DoF.

From a technical perspective, in~order to evaluate these variables, we need to take into consideration the angular movement caused in the pitch and the roll~axes.

First, regarding the pitch axis, the~movement is due to the longitudinal acceleration suffered in the~chassis, producing front and rear torsion on the springs and shock absorbers of the vehicle.
Given~the parameters $D_{pitch}$ and $K_{pitch}$, which represent the distance between the centre of the pitch axis with respect to the centre of mass of the vehicle and the characteristics of the spring together with the shock absorber, respectively,
Equation~(\ref{eq:pitchModel}) defines the dynamics of the pitch angle, which represents the sum of the moments of forces applied to the pitch axis.
The angular acceleration suffered by the vehicle chassis for the pitch axis is obtained by Equation~(\ref{eq:pitchModel2}), while the variables $\ddot{x}$, $\theta$, and~$\dot{\theta}$ are found in the LiDAR odometry process. 
\begin{equation}
(I_y + m\: d_{pitch}^2)\ddot{\theta} - m\: d_{pitch} \ddot{x} + (K_{pitch} + m\: g\: d_{pitch}) \theta + D_{pitch} \dot{\theta} = 0
\label{eq:pitchModel}
\end{equation}
\begin{equation}
\ddot{\theta} = -\frac{- m\: d_{pitch} \ddot{x} + (K_{pitch} + m\: g\: d_{pitch}) \theta + D_{pitch} \dot{\theta}}{I_y + m\: d_{pitch}^2}
\label{eq:pitchModel2}
\end{equation}
\noindent
where
\begin{equation}
\begin{aligned}
D_{pitch} &= \frac{d_{shock\: f}\ l_f^2 + d_{shock\: r}\ l_r^2}{2} \\
K_{pitch} &= \frac{K_{spring\: f}\ l_f^2 + K_{spring\: r}\ l_r^2}{2}
\end{aligned}
\end{equation}

With the pitch acceleration and applying Equation~(\ref{eq:pitchPredict}),
representing  uniformly accelerated motion, 
the pitch of the vehicle can be predicted at time $(t+1)$.
\begin{equation}
\widetilde{\theta}(t+1) = \frac{1}{2}\ddot{\theta}(t) dt^2 + \dot{\theta}(t) dt + \theta(t)
\label{eq:pitchPredict}
\end{equation}

On the other hand, the~angular movement caused on the roll axis is due to the lateral acceleration or lateral dynamics suffered in the chassis. The~parameter $d_{roll}$ is the distance between the roll axis centre and the centre of mass of the vehicle, and~mainly depends on the geometry of the suspension. The~lateral forces multiplied by the distance $d_{roll}$ generate an angular momentum, which is compensated for by the springs ($K_{rollf}, K_{rollr}$) and lateral shock absorbers of the vehicle ($D_{rollf}, D_{rollr}$), minimising the roll displacement suffered in the chassis. Equation~(\ref{eq:rollModel}) defines the movement dynamics of the roll angle, which represents the movement compensation effect with the sum of moments of forces applied on the axle.
\begin{equation}
(I_x + m\: d_{roll}^2)\ddot{\alpha} - m\: d_{roll} \ddot{y} + (K_{rollf} + K_{rollr} + m\: g\: d_{roll}) \alpha + (D_{rollf} + D_{rollr}) \dot{\alpha} = 0,
\label{eq:rollModel}
\end{equation}
\begin{equation}
\ddot{\alpha} = -\frac{- m\: d_{roll} \ddot{y} + (K_{rollf} + K_{rollr} + m\: g\: d_{roll}) \alpha + (D_{rollf} + D_{rollr}) \dot{\alpha}}{I_x + m\: d_{roll}^2},
\label{eq:rollModel2}
\end{equation}

\noindent
where
\begin{equation}
\begin{split}
D_{rollf} &= d_{shock\: f}\ t_f^2 \\
K_{rollf} &= \frac{K_{spring\: f}\ t_f^2}{2} \\
\end{split} ,
\hspace{2cm}
\begin{split}
D_{rollr} &= d_{shock\: r}\ t_r^2 \\
K_{rollr} &= \frac{K_{spring\: r}\ t_r^2}{2}
\end{split} .
\end{equation}

Given the roll acceleration and applying the uniformly accelerated motion Equation~(\ref{eq:rollPredict}), 
the roll of the vehicle can be predicted at time $(t+1)$:
\begin{equation}
\widetilde{\alpha}(t+1) = \frac{1}{2}\ddot{\alpha}(t) dt^2 + \dot{\alpha}(t) dt + \alpha(t).
\label{eq:rollPredict}
\end{equation}

Finally, to~complete the 6-DoF model parameterisation, we need to consider the vertical displacement of the vehicle, which is related to the angular movements of pitch and roll.
Equation~(\ref{eq:zModel}) represents the movement of the centre of masses concerning the vehicle z-axis, where $COG_z$ is the height of the vehicle's centre of gravity at resting state:
\begin{equation}
\widetilde{z}(t+1) = COG_z + d_{pitch} (\cos(\widetilde{\theta}(t+1)) - 1) + d_{roll} (\cos(\widetilde{\alpha}(t+1)) - 1)
\label{eq:zModel}
\end{equation}

Table~\ref{tab:parameters} lists the parameters and values used in the 6-DoF model. The~values correspond to a Volkswagen Passat B6, and~were found in the associated technical~specs.

\begin{table}[H]
\caption{Model parameters (chassis, tires, and~suspension).}
\centering
\begin{tabular}{cc}
\toprule
\textbf{Name} & \textbf{Value} \\
\midrule
$m$ = 1750 kg & Vehicle mass\\
$K_{spring\: f}$ = 30,800 $\frac{\text{N}}{\text{m}}$ & Front suspension spring stiffness\\
$K_{spring\: r}$ = 28,900 $\frac{\text{N}}{\text{m}}$ & Rear suspension spring stiffness\\
$D_{shock\: f}$ = 4500 $\frac{\text{Ns}}{\text{m}}$ & Front suspension shock absorber damping coefficient \\
$D_{shock\: r}$ = 3500 $\frac{\text{Ns}}{\text{m}}$ & Rear suspension shock absorber damping coefficient\\
$d_{roll}$ = 0.1 m & Vertical distance between COG and roll axis\\
$d_{pitch}$ = 0.25 m & Vertical distance between COG and pitch axis\\
$I_x$ = 540 kg m$^2$ & Vehicle's moment of inertia, with~respect to the x axis\\
$I_y$ = 2398 kg m$^2$ & Vehicle's moment of inertia, with~respect to the y axis\\
$I_z$ = 2875 kg m$^2$ & Vehicle's moment of inertia, with~respect to the  z axis\\
$COG_z$ = 0.543 m & COG height from the ground\\
$l_f$ = 1.07 m & Distance between COG and front axle\\
$l_r$ = 1.6 m & Distance between COG and rear axle\\
$t_f$ = 1.5 m & Front axle track width\\
$t_r$ = 1.5 m & Rear axle track width\\
\bottomrule
\end{tabular}
\label{tab:parameters}
\end{table}

To deal with the imperfections of the kinematic model, we compared the output of the proposed 6-DoF model with the ground truth available in the KITTI odometry data set. The~analysis was applied to all available sequences, in~order to measure the uncertainty model in the best~way.

\begin{figure}[H]
\centering
\includegraphics[width=0.9\linewidth]{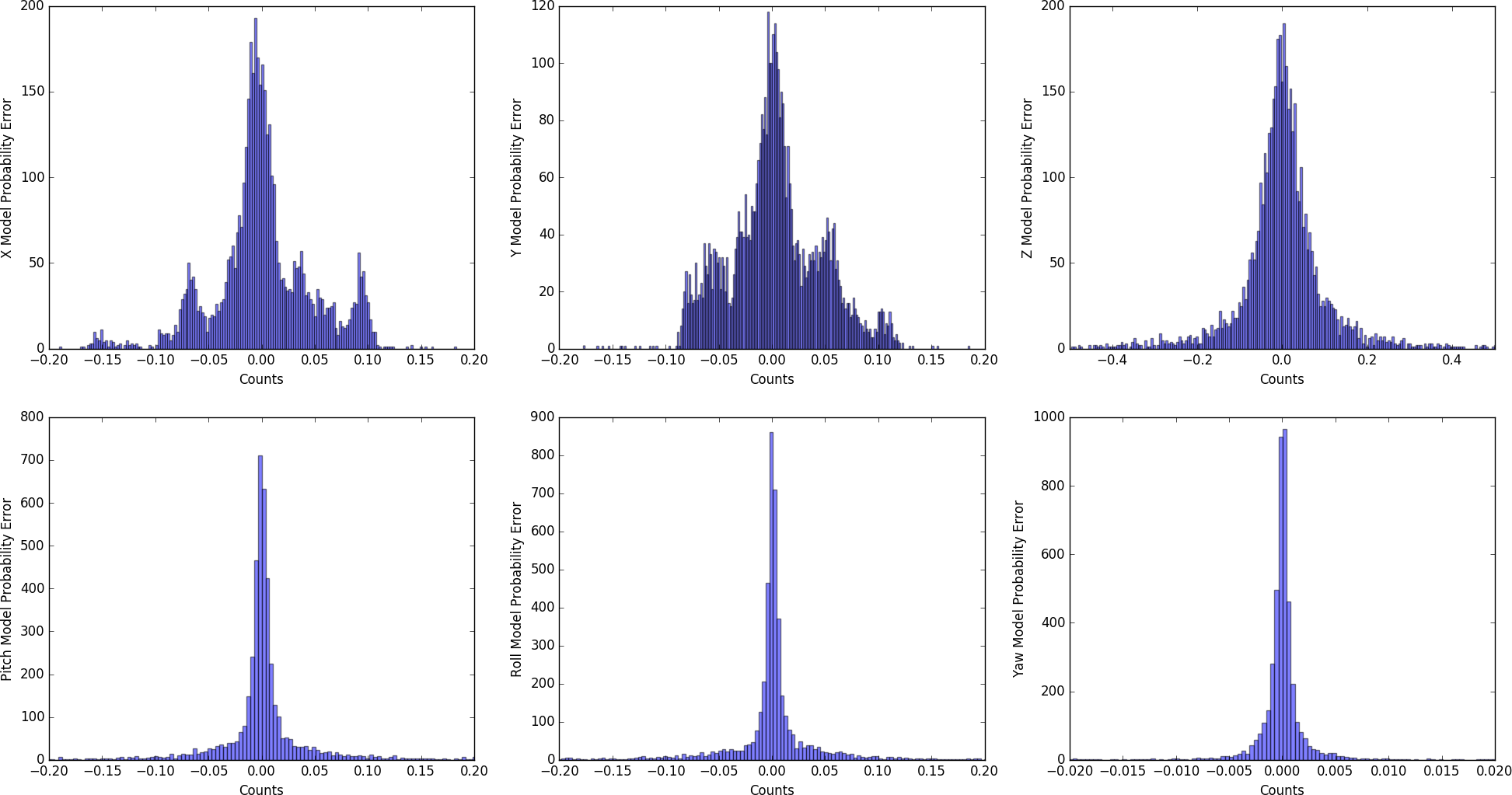}
\caption{Probabilistic error distribution representation for each vehicle output~variable.}
\label{fig:probabilisticDistribution}
\end{figure}

By evaluating the pose differences (see Figure~\ref{fig:probabilisticDistribution}), the~probability density function of the 6-DoF model was calculated, as~well as the covariance matrix expressed in Equation~(\ref{eq:covarianceMatrix}).
\begin{equation}
Q =
\begin{bmatrix}
{\sigma}_{xx}^2      & {\sigma}_{yx}^2      & {\sigma}_{zx}^2      & {\sigma}_{\phi x}^2      & {\sigma}_{\theta x}^2     & {\sigma}_{\psi x}^2 \\
{\sigma}_{xy}^2      & {\sigma}_{yy}^2      & {\sigma}_{zy}^2      & {\sigma}_{\phi y}^2      & {\sigma}_{\theta y}^2     & {\sigma}_{\psi y}^2 \\
{\sigma}_{xz}^2      & {\sigma}_{yz}^2      & {\sigma}_{zz}^2      & {\sigma}_{\phi z}^2      & {\sigma}_{\theta z}^2     & {\sigma}_{\psi z}^2 \\
{\sigma}_{x\phi}^2   & {\sigma}_{y\phi}^2   & {\sigma}_{z\phi}^2   & {\sigma}_{\phi\phi}^2    & {\sigma}_{\theta\phi}^2   & {\sigma}_{\psi\phi}^2 \\
{\sigma}_{x\theta}^2 & {\sigma}_{y\theta}^2 & {\sigma}_{z\theta}^2 & {\sigma}_{\phi\theta}^2  & {\sigma}_{\theta\theta}^2 & {\sigma}_{\psi\theta}^2 \\
{\sigma}_{x\psi}^2   & {\sigma}_{y\psi}^2   & {\sigma}_{z\psi}^2   & {\sigma}_{\phi\psi}^2    & {\sigma}_{\theta\psi}^2   & {\sigma}_{\psi\psi}^2
\end{bmatrix} ,
\label{eq:covarianceMatrix}
\end{equation}
where ${\sigma}_{xx}$ = 0.0485 m, ${\sigma}_{yy}$ = 0.0435 m, ${\sigma}_{zz}$ = 0.121 m, ${\sigma}_{\phi\phi}$ = 0.1456 rad, ${\sigma}_{\theta\theta}$ = 0.1456 rad, ${\sigma}_{\psi\psi}$ = 0.0044~rad, and~the error covariance between variables has a zero~value.

\unskip

\section{Vehicle Pose Estimation~System}
\label{sec:filteringUKF}

\begin{figure}[H]
\centering
\includegraphics[width = 0.45\linewidth] {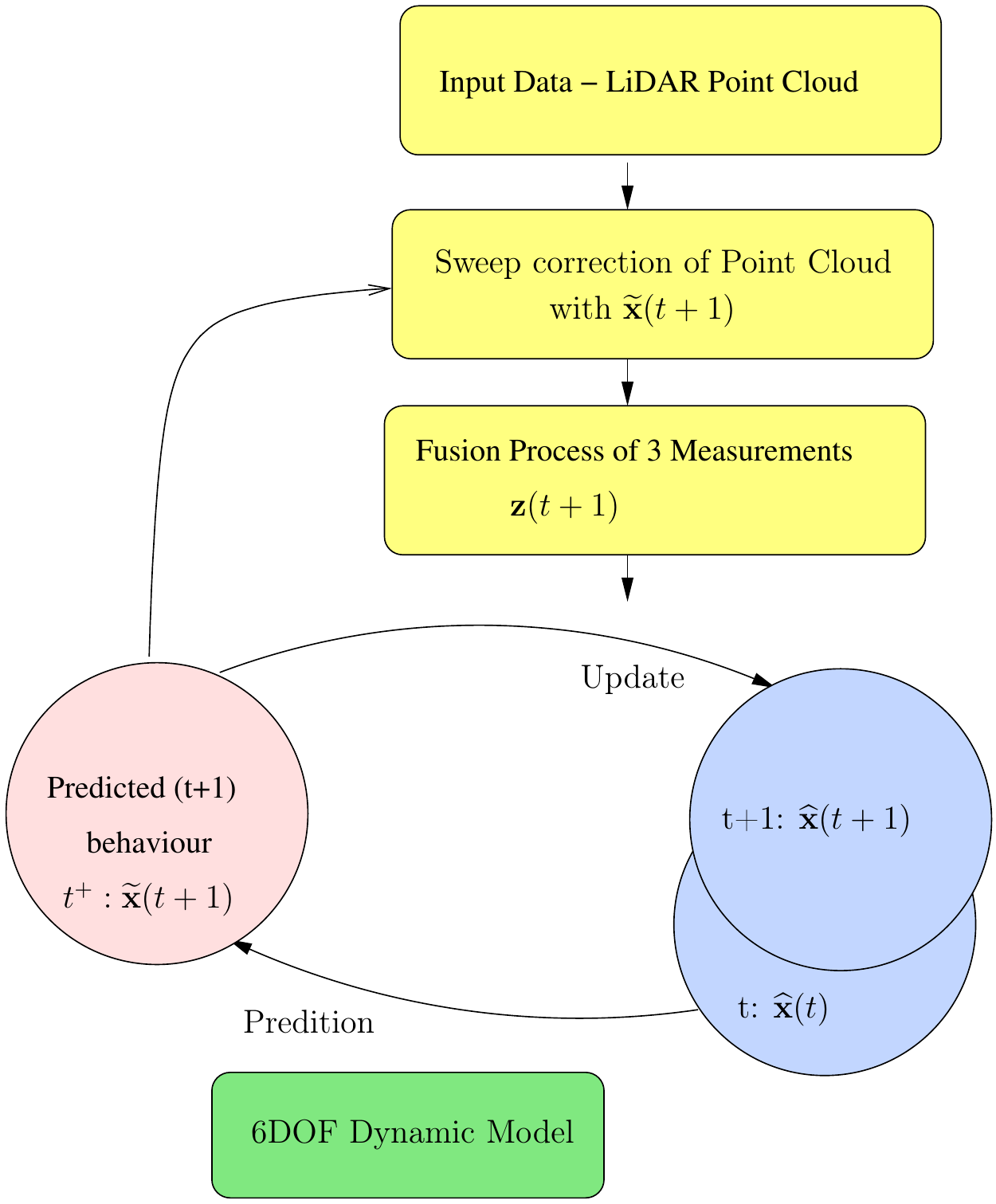}
\caption{Unscented Kalman Filter (UKF) architecture. The~prediction phase relies on the 6-DoF motion model detailed in the previous section. The~update phase uses three consecutive LiDAR-based measurements to fuse and estimate the vehicle~state.}
\label{fig:kalmanDiagram}
\end{figure}

This section details the architecture implemented to estimate the vehicle's attitude, by~integrating the dynamic and kinematic model described in Section~\ref{sec:model} and fusing the LiDAR-based measurement system described in Section~\ref{sec:measurements}. Several works have analysed the response of two of the most well-known filters for non-linear models, the~Extended Kalman Filter (EKF) and the Unscented Kalman Filter (UKF), where the results were generally in favour of the UKF. For~instance, in~\cite{DALFONSO2015122}, the~behaviour of both filters was compared to estimate the position and orientation of a mobile robot. Real-life experiments showed that UKF has better results in terms of localisation accuracy and consistency of estimation. The~proposed architecture therefore integrates an Unscented Kalman Filter~\cite{eric}, which is divided into two stages: prediction and update (as shown in Figure~\ref{fig:kalmanDiagram}).

\unskip

The prediction phase manages the 6-DoF dynamic model described in the previous section to predict the system's state at time $(t+1)$. Along with the definition of the model, the~model noise covariance matrix $Q$ must be associated, as~defined by the standard deviations evaluated above. The~model noise covariance matrix is only defined in its main diagonal and is constant over time. Equation~(\ref{eq:prediction}) represents the prediction phase of the filter.
\begin{equation}
\mathbf{\widetilde{x}}(t+1) = A \mathbf{\widehat{x}}(t) + Q,
\label{eq:prediction}
\end{equation}
where $\mathbf{x}$ is the 6-DoF state vector, as~shown in Equation~(\ref{eq:stateVectorDefinition}), and~$A$ matrix represents the  developed dynamic model.
\begin{equation}
\mathbf{x}'(t) =
\begin{bmatrix}
{x}(t) \; {y}(t) \; {z}(t) \; \alpha(t) \; \theta(t) \; \psi(t)
\end{bmatrix}.
\label{eq:stateVectorDefinition}
\end{equation}

In the filter update phase, the~LiDAR odometry output is estimated. The~estimated state vector, $\widehat{\mathbf{x}}(t+1)$, is represented, in~terms of the state variables, by~Equation~(\ref{eq:update}). The~6 $\times$ 6 matrix $C$ is defined with the identity matrix, as~the vectors $\mathbf{z}(t+1)$ and $\widehat{\mathbf{x}}(t+1)$ contain the same measurement units. Finally, the~matrix $R$ is the covariance error matrix of the measurement, which is updated every odometry period in the measurement and fusion process, as~explained in Section~\ref{sec:measurements}. The~matrix $R$ is only defined in its main diagonal, representing the uncertainty of each of the magnitudes measured in the process.
\begin{equation}
\mathbf{z}(t+1) = C \mathbf{\widehat{x}}(t+1) + R .
\label{eq:update}
\end{equation}

\subsection*{LiDAR Sweep~Correction}

\begin{figure}[H]
\centering
\begin{subfigure}[]{.5\linewidth} \centering \includegraphics[width = 0.9\linewidth] {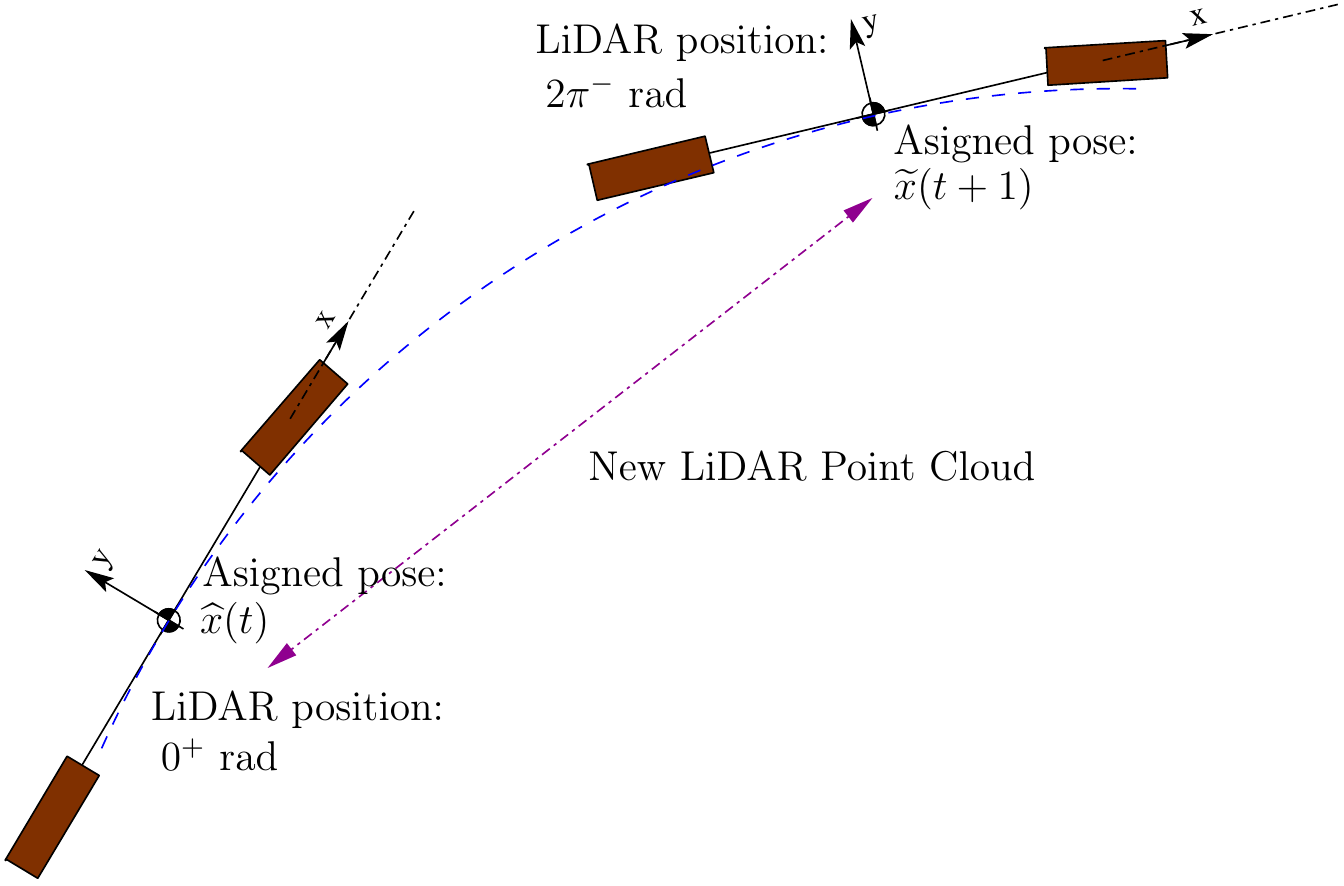} \caption{} \end{subfigure}
\begin{subfigure}[]{.4\linewidth} \centering \includegraphics[width = 0.9\linewidth] {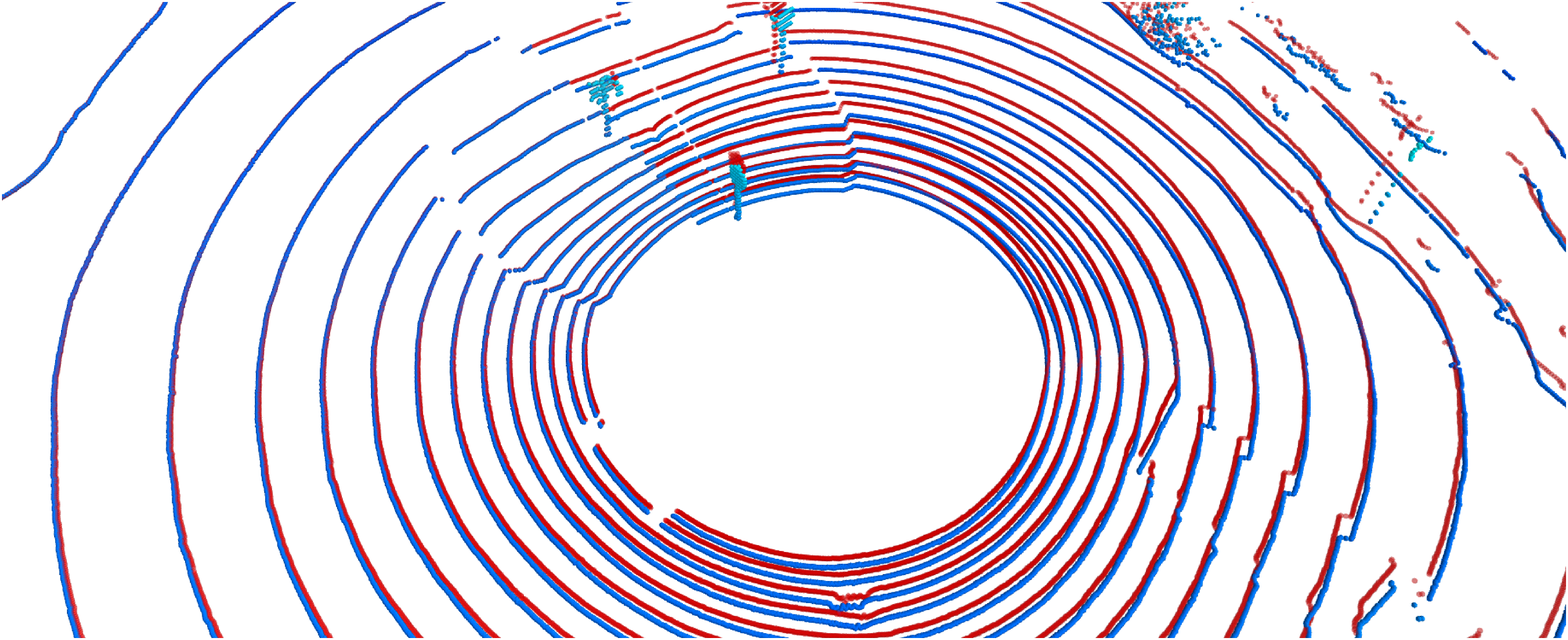} \caption{} \end{subfigure}
\caption{Sweep correction process in odometry: (\textbf{a}) the assignment of two poses to the point cloud when the vehicle is moving; and (\textbf{b}) raw (blue) and corrected measurements (red). An~important difference between the measurement results of both point clouds is exposed, the~correction being decisive for the result of the following~stages.}
\label{fig:sweepCorrection}
\end{figure}

To use the LiDAR data in the update phase of the UKF, it is recommended to perform a so-called {sweep correction} of the raw data. The~sweep correction phase is due to the nature of most LiDAR devices, which are composed of a series of laser emitters mounted on a spinning head (e.g., the~Velodyne HDL-64E).
The sweep correction process becomes crucial when the sensor is mounted on a moving vehicle, as~the sensor spin requires a time span close to approximately  100~ms, as~in the case of the Velodyne HDL-64E. The~sweep correction process consists of assigning two poses for each sensor output and interpolates the poses with constant angular speed for all the LiDAR beams. These poses are commonly associated with the beginning and the end of the sweep. Thus, the~initial pose is equal to the last filter estimation $\widehat{\mathbf{x}}$ and the final pose is equal to the filter prediction $\widetilde{\mathbf{x}}(t+1)$ to carry out the interpolation. The~whole point cloud is corrected with the interpolated poses evaluated, solving the scene deformation issue when the LiDAR sensor is mounted on a moving platform. Figure~\ref{fig:sweepCorrection}a shows the key points on the sweep correction~process.

Regarding the correction method, the~authors in~\cite{DBLP:journals/corr/MerriauxDBVS17, B_Zhang} proposed a point cloud correction procedure based on GPS data. The~process requires synchronisation between each GPS and LiDAR output, a~complex task when the GPS introduces small delays in its measurement. For~this reason, in~our case, the~GPS data is replaced with the filter prediction to apply the sweep correction process.
Figure~\ref{fig:sweepCorrection}b shows the same point cloud with and without sweep correction, captured in a roundabout with low angular speed vehicle movement. It can be seen that there is significant distortion concerning reality, as~the difference of shapes between clouds is substantial, leading to errors of one meter in many of the scene elements. We can claim that the motion model accuracy is a determinant for the sweep correction process, as~it improves the odometry results (as we depict in Section~\ref{section:experimentalAnalysis}).

\unskip

\section{Measurements Algorithms and Data~Fusion}
\label{sec:measurements}

Three measurement methodologies based on LiDAR raw-data were developed to provide an accurate and robust algorithm. Two of them are based on ICP techniques, and~another one relies on feature extraction and SVD. A~6-DoF measure $z(t)$ is the output of this process, after~the fusion process is~finished.

\subsection{Multiplex General~ICP}

Using the ICP algorithm for the development of LiDAR odometry systems is very common, where the two most used versions are the Point-to-Point and Point-to-Plane schemes. Adaptations of both algorithms have been developed for our approach.  For~the first measurement system developed, we propose the use of the ICP point-to-point algorithm, which is based on aligning two partially overlapping point clouds to find the homogeneous transformation matrix $(R, t)$ in order to align the two point clouds. The~ICP used is based on minimising the cost function defined by the mean square error of the distances between points in both clouds, as~expressed in Equation~(\ref{eq:icp}). The~point cloud registration follows the criterion of the nearest neighbour distance between clouds.
\begin{equation}
\label{eq:icp}
\underset{R,T}{min}(error(R,T))=\underset{R,T}{min}(\frac{1}{N_p}\sum_{1}^{N_p}{\left \| p_i - (q_iR+T) \right \|}),
\end{equation}
where $p_i$ represents the set of points that defines the cloud captured at a time instant $(t-1)$, $q_i$~represents the set of points that define the cloud captured at a time instant $t$, $N_p$ is the number of points considered in the minimisation process, $R$ is the resulting rotation matrix, and~$T$ is the resulting translation~matrix.

\begin{figure}[H]
\centering
\includegraphics[width = 0.8\linewidth]{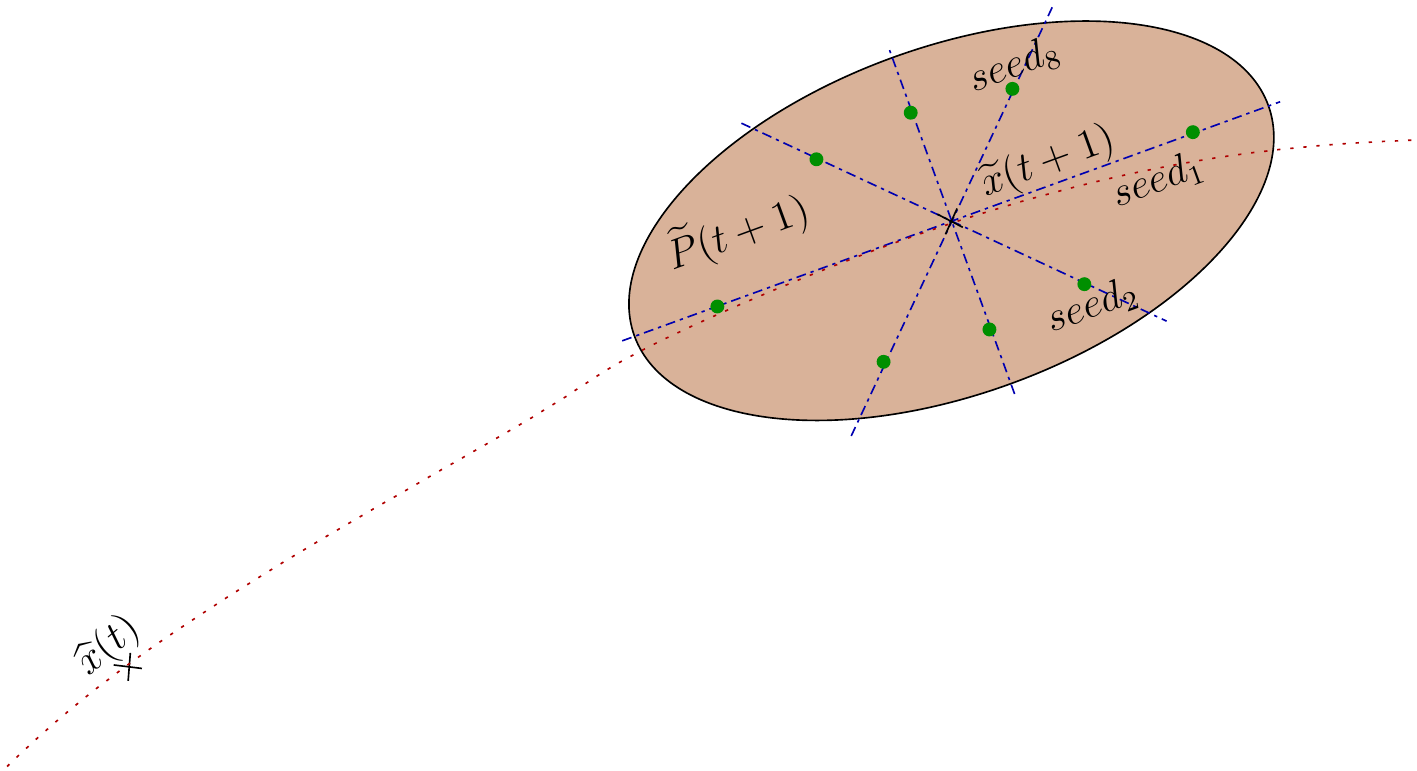}
\caption{Sharing of Iterative Closest Points  (ICP) initial conditions applying the Sigma Point techniques in the limits marked by $\widetilde{P}(t+1)$.}
\label{fig:seed_ICP}
\end{figure}

The ICP technique, as~with many other gradient descent methods, can become stuck at a local minimum instead of the global minimum, causing measurement errors. The~possibility of finding moving objects or a lack of features in the scene are some of the reasons why the ICP algorithm provides local minimum solutions. For~this reason, an~algorithm that computes the multiplex ICP algorithm for a set of distributed seeds was implemented. The~selected seed, such as the ICP starting point, is evaluated with the Merwe Sigma Points method~\cite{Merwe,Julier}. The~error covariance matrix predicted by the filter $\widetilde{P}(t+1)$ and the predicted state vector $\widetilde{x}(t+1)$ are the input to assess the eight seeds needed. Figure~\ref{fig:seed_ICP} shows an example of seed distribution in the plane $(x,y)$ for a time instant $(t)$.

After evaluating the eight measures, the~one with the best mean square error in the ICP process is selected. After~the evaluation of two sequences of the KITTI data set, a~decrease of error close to 9.5\% was discerned, this being the determining reason why eight seeds were selected. However, the~increase in computation time could be a~disadvantage.

\subsection{Normal Filtering~ICP}

For the design of a robust system, it is not enough to integrate only one measurement technique, as~it may fail due to multiple factors. Therefore, a~second measurement method based on ICP point-to-plane was developed to improve the robustness of the system, as~it implies a lower computation time than the one above. In~\cite{chen_medioni}, the~results with the point-to-plane method were more precise than those with the point-to-point method, improving the precision of the measurement. The~cost function to be minimised in the point-to-plane process is as follows (\ref{eq:icpNormalFiltering}):
\begin{equation}
\label{eq:icpNormalFiltering}
\underset{R,T}{min}(error(R,T))=\underset{R,T}{min}(\frac{1}{N_p}\sum_{1}^{N_p}{\left \| (p_i - (q_iR+T)) \cdot n_i^p \right \|}),
\end{equation}
where $R$ and $T$ are the rotation and translation matrices, respectively, $N_p$ is the number of points used to optimize, $p_i$ represents the source cloud, $q_i$ represents the target cloud, and~$n_i^p$ represents the normal unit vector of a point in the target cloud. The~point-to-plane technique is based on a weighting to register cloud points in the minimisation process, where $cos(\theta)$ from the vectorial product is the weight given in the process and $\theta$ is the angle between the unit normal vector $n_i^p$ and the vector resulting from the operation $(p_i - (q_iR+T))$. Therefore, the~smaller the angle $\theta$ is, the~higher the contribution in the added term of this register point is. So, the~normal unit vector $n_i^p$ can be understood as rejecting or decreasing the impact over the added term of its register points when the alignment with the unitary vector is not right. The~approach in this paper does not include all the points registered, as~a filter process is carried out. The~heading of the vehicle is the criteria to implement the filtering process. Thus, only those points that have a normal vector within the range $\widetilde{\psi}_v \pm \widetilde{\sigma}_{\psi\psi} rad$ are considered in the added term, where $\widetilde{\psi}_v$ represents the heading of the predicted vehicle and $\widetilde{\sigma}_{\psi\psi}$ represents the uncertainty predicted from the error covariance matrix. Equation~(\ref{eq:normalCriterium}) formulates the criteria applied in the minimisation to filter out points:
\begin{equation}
\label{eq:normalCriterium}
\begin{matrix}
min J(R,T) & = & \underset{R,T}{min}(\frac{1}{N_p}\sum_{1}^{N_p}{\left \| (p_i - (q_iR+T)) \cdot n_i^p \right \|})\\
s.t. & & \\
n_i^p > \widetilde{\psi}_v \pm \widetilde{\sigma}_{\psi\psi} & & \\
n_i^p > \widetilde{\psi}_v \pm \widetilde{\sigma}_{\psi\psi} + \pi & & \\
\end{matrix}
\end{equation}

\begin{figure}[H]
\centering
\begin{subfigure}[]{.46\linewidth} \centering  \includegraphics[width = 0.9\linewidth] {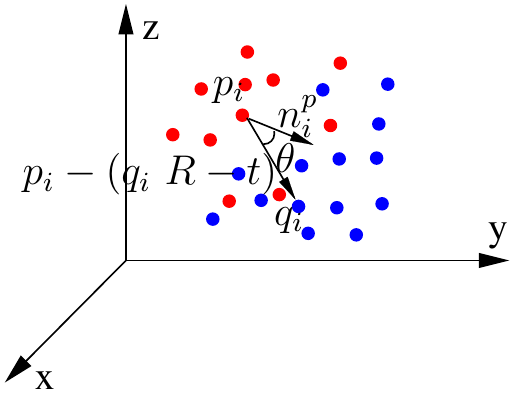} \caption{} \end{subfigure}
\begin{subfigure}[]{.46\linewidth} \centering  \includegraphics[width = 0.9\linewidth] {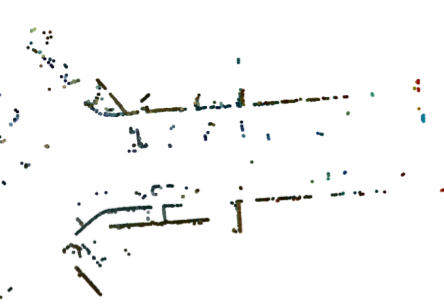} \caption{} \end{subfigure}

\caption{ICP process based on normals: (\textbf{a}) Graphical representation of the cost function with the normal unit vector $n_i^p$ used to enter constraints in ICP; and (\textbf{b}) ICP output result applying constraints of normals. The~figure shows the overlap of two consecutive~clouds.}
\label{fig:rejectingPoints}
\end{figure}

Points that are not aligned with the longitudinal and transverse directions of the vehicle are eliminated from the process, improving the calculation time of this process as well as the accuracy of the measurement. Figure~\ref{fig:rejectingPoints} represents an ICP iteration of the described technique, where the results achieved by RMSE are 20\% better than if all the points of the cloud are~considered.

\unskip

\subsection{SVD Cornering~Algorithm}

The two previous systems of measurement are ICP-based techniques, where there is no known data association between the points of two consecutive point clouds. However, the~third~proposed algorithm uses synthetic points generated by the algorithm and the data association of the synthetic point between point clouds to evaluate the odometry step. An~algorithm for extracting the characteristics within the point clouds is developed to assess the synthetic points. The~corners built up with the intersection between planes are the features explored. The~SVD algorithm uses the corners detected in consecutive instants to determine the odometry between point clouds. The~new odometry complements the two previous measurements. The~SVD algorithm is accurate and has low computational load, although~the computation time increases in the detection and feature extraction~steps.

\subsubsection{Synthetic Point~Evaluation}
\unskip
\paragraph{Plane Extraction}

\begin{figure}[H]
\centering
\begin{subfigure}[]{.4\linewidth} \centering \includegraphics[width = 0.9\linewidth]{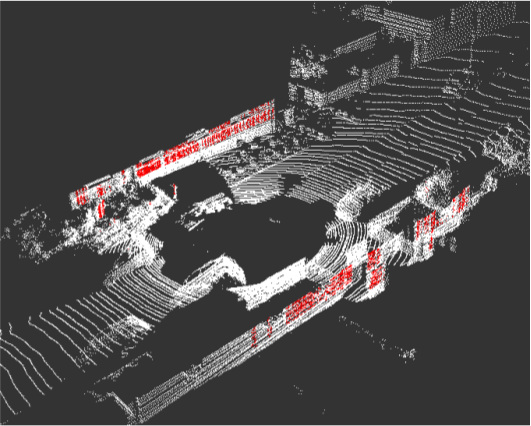} \caption{} \end{subfigure}
\begin{subfigure}[]{.375\linewidth} \centering \includegraphics[width = 0.9\linewidth]{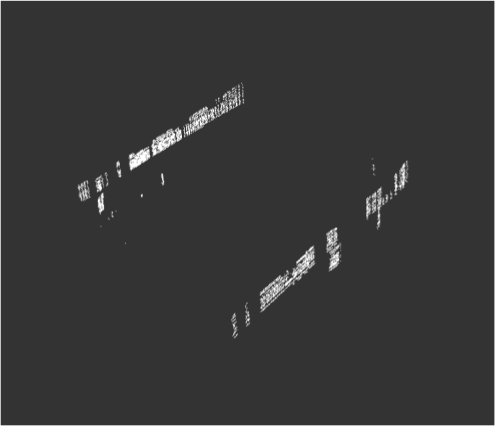} \caption{} \end{subfigure}
\caption{Intermediate results of sequence 00, frame 482: (\textbf{a}) Input cloud to the plane detection algorithm; and (\textbf{b}) points detected on candidate~planes.}
\label{fig:planeDetection1}
\end{figure}

It is easy for humans to identify flat objects in an urban environment; for instance, building~walls. However, identifying vertical planes in a point cloud with an algorithm is more complex. The~algorithm identifies points that, at~random heights, fit the same projection on the plane $(x,y)$. Therefore, the~number of repetitions that each beam of the LiDAR presents on the plane $(x,y)$ is recorded. If~the number of repetitions of the project exceeds the threshold of 20 counts, the~points belong to a vertical plane. Figure~\ref{fig:planeDetection1} shows detected points that belong to vertical planes, although~the planes in many cases are not~segmented.

\paragraph{Clustering}

\begin{figure}[H]
\centering
\begin{subfigure}[]{.46\linewidth} \centering  \includegraphics[width = 0.9\linewidth] {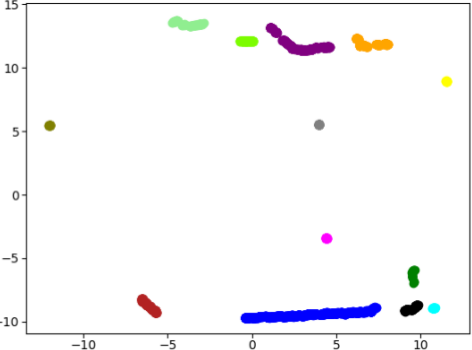} \caption{} \end{subfigure}
\begin{subfigure}[]{.46\linewidth} \centering  \includegraphics[width = 0.9\linewidth] {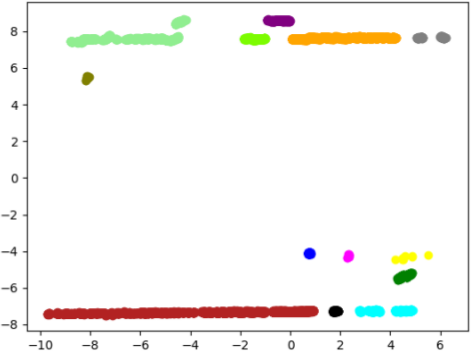} \caption{} \end{subfigure}
\caption{Results of clustering, sequence 00 of the KITTI odometry data set: (\textbf{a}) Frame 0, and~({\textbf{b}})~Frame~482.}
\label{fig:ClusteringDBSCAN}
\end{figure}

Clustering techniques are then used to group the previously selected points into sets of intersecting planes. Among~those listed in the state-of-the-art, those that do not imply knowing the number of clusters to be segmented were considered valid, as~it is not known a priori. Analysing the clustering results provided by the \textit{Sklearn} library, DBSCAN was the one that obtained the best results, as~it does not make mistakes when grouping points of the same plane in different clusters. In~order to provide satisfactory results, the~proposed configuration of the DBSCAN clustering algorithm sets the maximum distance between two neighbouring points (0.7) and the minimum number of samples between neighbours (50). The~algorithm identifies solid structure corners, such as building walls, such that clusters associated to non-relevant structures are eliminated. For~this purpose, clusters sized smaller than 300 points were filtered, eliminating noise produced by vegetation or pedestrians. Figure~\ref{fig:ClusteringDBSCAN} represents the cluster segmentation of the point cloud depicted in Figure~\ref{fig:planeDetection1}, where only the walls of buildings, street lights, or~traffic signs are segmented as characteristic elements of the~scene.

\paragraph{Corner Detection and Parameters Extraction}

\begin{figure}[H]
\centering
\begin{subfigure}[]{.46\linewidth} \centering  \includegraphics[width = 0.9\linewidth] {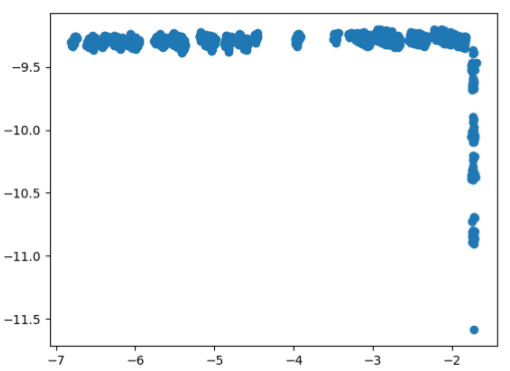} \caption{} \end{subfigure}
\begin{subfigure}[]{.46\linewidth} \centering  \includegraphics[width = 0.9\linewidth] {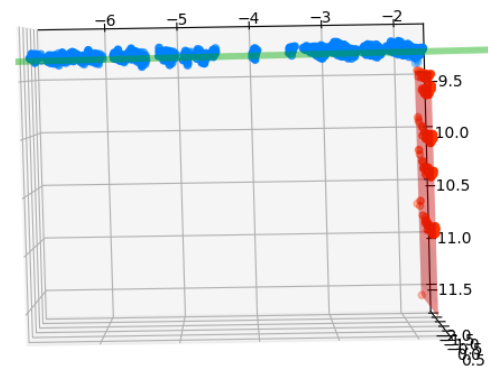} \caption{} \end{subfigure}
\caption{Results of extraction of intersected planes: (\textbf{a}) Input data to cluster planes; and (\textbf{b}) detection results of two intersecting planes, represented in red and~green.}
\label{fig:twoSecantPlanes}
\end{figure}

In addition, to~eliminate straight walls, cylindrical points, or~a variety of shapes that are not valid for the development of the algorithm, clusters that do not contain two vertical intersected planes are discarded, as~shown in Figure~\ref{fig:ClusteringDBSCAN}. Thus, two intersecting planes are searched for in the cluster that satisfies the condition of forming an angle between both higher than $45^{\circ}$ and less than $135^{\circ}$. Using the RANSAC algorithm on the complete set of points of the cluster, indicating that it selects a quarter of the total points and fixing the maximum number of iterations as 500 iterations, the~algorithm returns the equation of a possible intersected plane in the cluster. Applying RANSAC again to the outlier points resulting from the first process and with the same configuration parameters, a~second intersected plane in the cluster is achieved, as~shown in Figure~\ref{fig:twoSecantPlanes}. If~the angle formed between the two intersected planes fulfils the previous conditions, the~intersection line of both planes is evaluated to obtain the synthetic points that define the evaluated~corner.

\paragraph{Synthetic Points Evaluation}

\begin{figure}[H]
\centering
\begin{subfigure}[]{.32\linewidth} \centering \includegraphics[width = 0.9\linewidth] {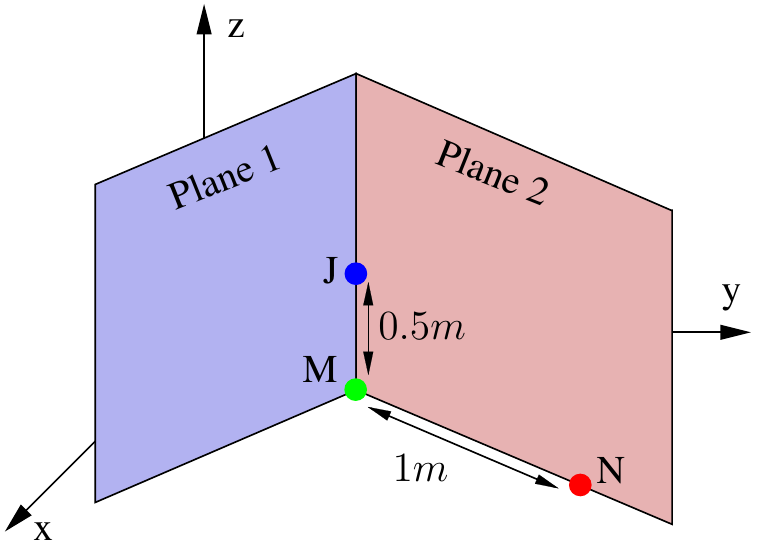} \caption{} \end{subfigure}
\begin{subfigure}[]{.65\linewidth} \centering \includegraphics[width = 0.9\linewidth] {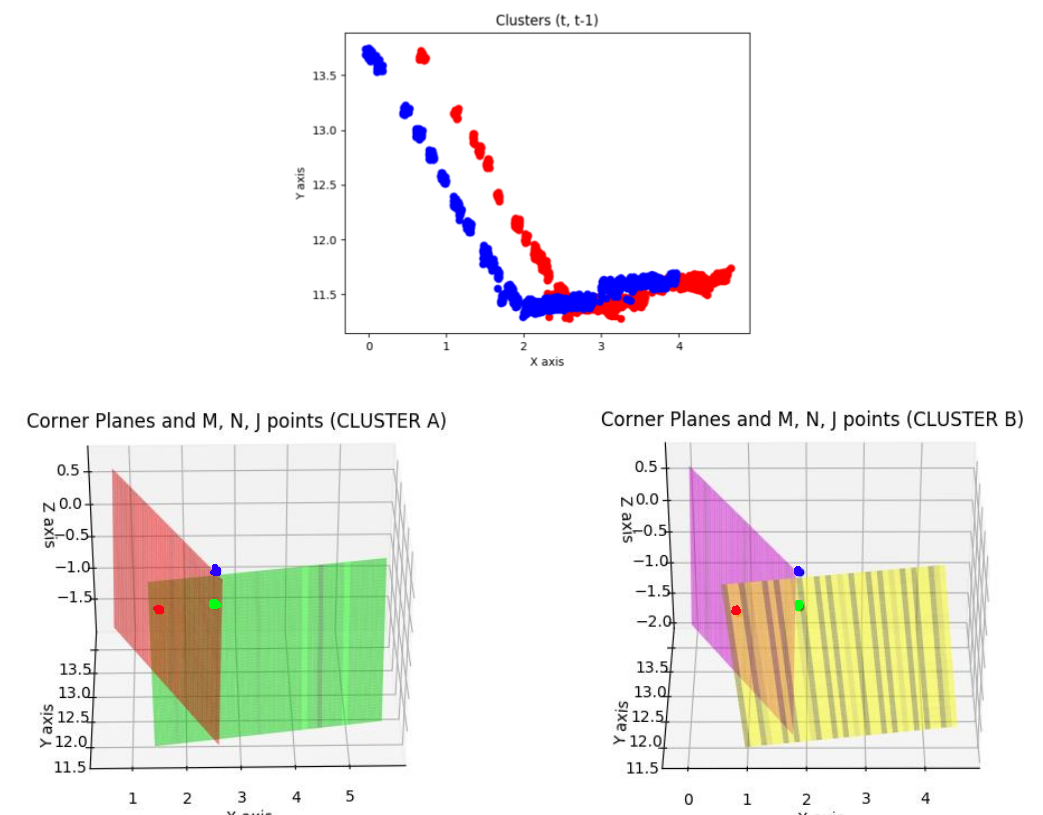} \caption{} \end{subfigure}
\caption{Evaluation of synthetic points: (\textbf{a}) Nomenclature and position of calculated synthetic points; and (\textbf{b}) result of synthetic points detection in real clusters of two consecutive time~instants.}
\label{fig:synteticPoints}
\end{figure}

At this point, the~objective is to generate three points that characterize the corners of the scene; these points are denoted as synthetic points. The~synthetic points are obtained from the intersection line equation derived from the two intersecting planes. Figure~\ref{fig:synteticPoints}a shows the criterion followed to evaluate three synthetic points for each of the detected corners. Two of the synthetic points, $(M, J)$, belong to the intersection line and are located at a distance of 0.5~m. The~third synthetic point, $N$, meets the criterion of being at 1 m of point separation from $M$ with a value of $z = 0$. The~process identifies, as~the reference plane, the~one that has the lowest longitudinal plane direction evaluated within the global co-ordinate system. Figure~\ref{fig:synteticPoints}b shows the points $(M, J, N)$ evaluated in two consecutive instants of time. In~this situation, the~SVD algorithm can be applied to assess the homogenous transform between two consecutive point clouds when the synthetic points data association is~known.

\subsubsection{SVD}

Before applying the SVD, the~registration of the extracted points of the corners between two~consecutive instants need to be done. So, let us suppose that, for~an instant $t$, there is a set of corners $X={x_1, x_2, x_3, ..., x_n}$ where $x_1 = (M_1, N_1, J_1)$ and, for~another instant $t+1$, there is another set of corners $Y={y_1, y_2, y_3, . ...y_m}$ where $y_1 = (M_1, N_1, J_1)$. Then, to~register both sets, the~Euclidean distance of  points $M$ is used. Only those corners that show a minimum distance less than 0.5 m are data associated. The~non-data associated corners are~removed.

Once the data association of synthetic corners is fulfilled, the~objective is to find the homogeneous transformation between two consecutive scenes. Therefore, SVD minimises the distance error between synthetic points, first by eliminating the translation of both sets to exclude the unknown translation and then by solving the Procustes orthogonal problem to obtain the rotation matrix $(R)$. Finally, it~undoes the translation to obtain the translation matrix $(T)$. Equation~(\ref{eq:SVDDescription}), described in more detail in~\cite{wu2018fast}, shows the mathematical expressions applied in the SVD algorithm to obtain the homogeneous transformation matrix between two sets of synthetic corners at consecutive time instants:
\begin{equation}
\label{eq:SVDDescription}
\begin{matrix}
{X}' = {x_i - \mu_x} = {{x_i}'} \\
{Y}' = {y_i - \mu_y} = {{y_i}'} \\
W = \sum_{i=1}^{N_p}{x_i}'{y_i}'^T\\
W = U\sum V^T\\
R = UV^T\\
t = \mu_x - R\mu_y
\end{matrix} .
\end{equation}

The SVD odometry measure $z_{SVD}$ is fused with the other measurements, but~the factor related to the uncertainty must be added to the SVD homogeneous transformation $\Delta\: Pose_{SVD}$. Therefore, Equation~(\ref{eq:zSVD}) defines the SVD measure added to the $\Delta\: Pose_{SVD}$, the~UKF estimated state vector $\widetilde{x}(t)$, and~the uncertainty factor $R_{SVD}$. The~uncertainty represents the noise covariance matrix of the SVD measurement and $R_{SVD}$ is calculated with the RMSE returned by the RANSAC process applied within the method. The~decision taken is a consequence of distinguishing a direct relationship between $R_{SVD}$ and how well the intersection planes are fitted over the points of the cluster.
\begin{equation}
z_{SVD} = \widetilde{x}(t) + \Delta\: Pose_{SVD} + R_{SVD} .
\label{eq:zSVD}
\end{equation}

\begin{figure}[H]
\centering
\includegraphics[width = 0.85\linewidth] {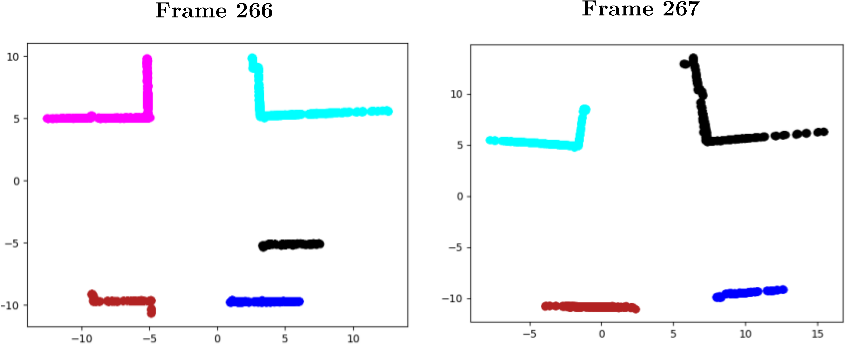}
\caption{Odometry results with Singular Value Decomposition (SVD): Input cluster to extract synthetic~points.}
\label{fig:resultsSVD}
\end{figure}

\begin{figure}[H]
\centering
\begin{subfigure}[]{.45\linewidth} \centering  \includegraphics[width = 0.9\linewidth] {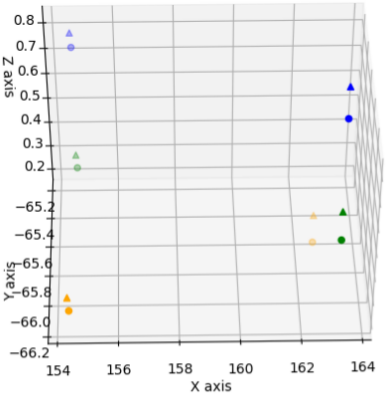} \caption{} \end{subfigure}
\begin{subfigure}[]{.45\linewidth} \centering  \includegraphics[width = 0.9\linewidth] {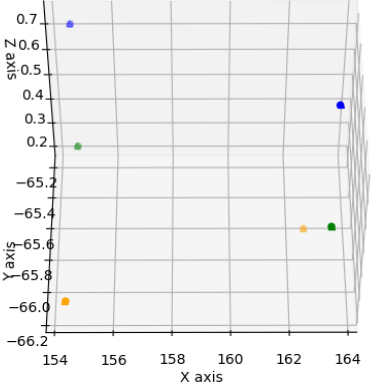} \caption{} \end{subfigure}
\caption{Odometry results with SVD: (\textbf{a}) Representation of the synthetic points extracted from the previous clusters. A~translation and rotation between them is shown; and (\textbf{b}) synthetic points are overlapped when applying the rotation and translation calculated by~SVD.}
\label{fig:resultsSVD1}
\end{figure}

Figures~\ref{fig:resultsSVD} and \ref{fig:resultsSVD1} depict a successful scenario where SVD odometry is evaluated. The~colour code used in the figure is: green (M points), blue (J points), and~orange (N points).

\unskip

\unskip

\subsection{Fusion~Algorithm}

\begin{figure}[H]
\centering
\begin{subfigure}[]{.35\linewidth} \centering \includegraphics[width = 0.9\linewidth] {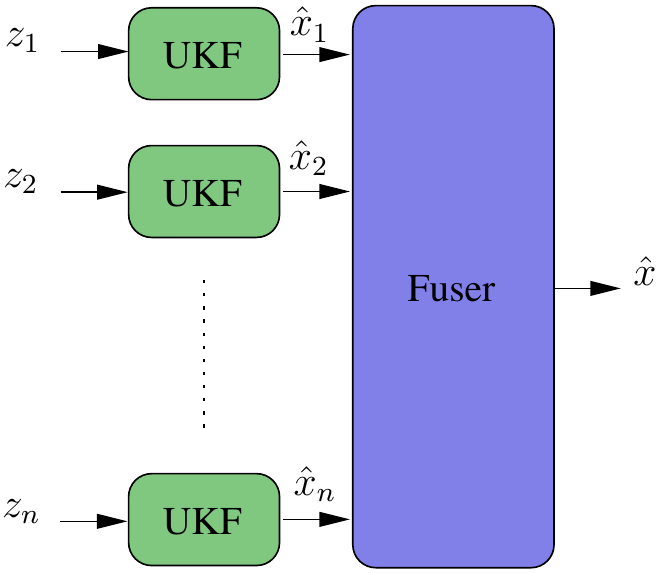} \caption{} \end{subfigure}
\begin{subfigure}[]{.35\linewidth} \centering \includegraphics[width = 0.9\linewidth] {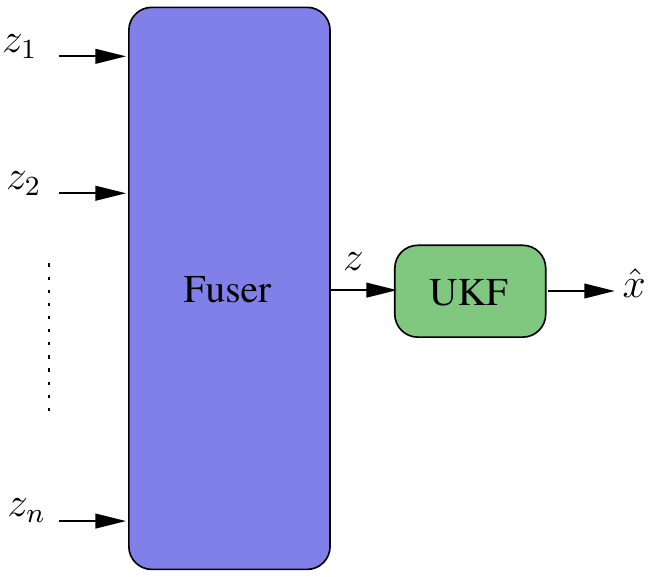} \caption{} \end{subfigure}
\caption{Block diagram for two fusion philosophies: (\textbf{a}) Merging of the estimated state vector, which~requires a filtering stage for each measure to be merged; and (\textbf{b}) merging of observations under a given criterion and subsequent~filtering.}
\label{fig:fusionKalman}
\end{figure}

An essential attribute in the design of a robust system is the redundancy. For~the proposed~work, three measurement techniques based on LiDAR were developed.  Therefore, it is necessary to integrate sensor fusion techniques that allow for selecting and evaluating the optimal measurement from an input set to the solution. Figure~\ref{fig:fusionKalman}a shows an architecture where the filter outputs---that is, the~estimated state vectors---are fused. The~main architecture characteristic is that multiplex filters have to be integrated into the solution. Figure~\ref{fig:fusionKalman}b shows an architecture that fuses a set of measurements and then filters the fused measurement. For~this architecture design, only blocks have to be designed, improving its simplicity. In~this second case, all the measurements must represent the same magnitude to be measured. In~\cite{913685}, a~system that merges the data from multiple sensors using the second approach was presented. The~proposed fusion system implements this sensor fusion architecture, in~which the resulting measurement vector comprises the 6-DoF of the~vehicle.

\begin{figure}[H]
\centering
\includegraphics[width = 0.75\linewidth] {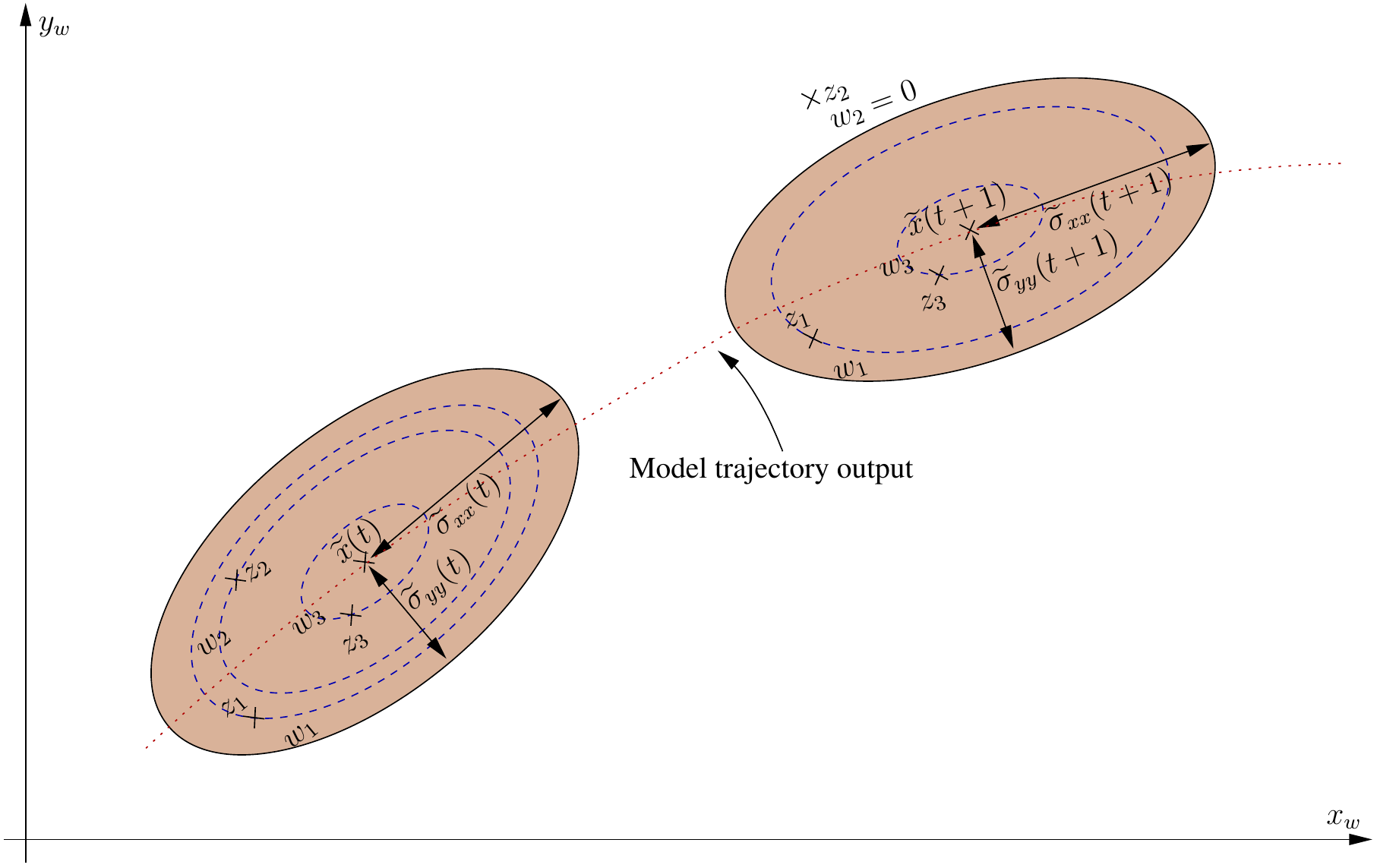}
\caption{Representation of predicted ellipses of uncertainty and weight allocation to each measure to be applied in~fusion.}
\label{fig:pesosMedidas}
\end{figure}

The proposed sensor fusion consists of assigning a weight to each of the measurements. The~weights are evaluated considering the distance (x, y) between the filter prediction and the LiDAR-based measurements. Therefore, Equation~(\ref{eq:ecuacionPonderacion}) defines the weighting function. The~assigned weight varies between 0 and 1 when the measurement is within the uncertainty ellipse. The~assigned weight is 0 when the measurement is outside the uncertainty ellipse, as~shown in Figure~\ref{fig:pesosMedidas}. The~predicted error covariance matrix $\widetilde{P}(t+1)$ defines the uncertainty ellipse. The~weighted mean value is the fused measurement, as~detailed in Equation~(\ref{eq:valorMedio}). In~the same way, the~uncertainty associated with the fused measure is weighted with the partial measure weight. Thus, the~sensor fusion output is a 6-DoF measure with an associated uncertainty matrix $R$.

\begin{equation}
\label{eq:ecuacionPonderacion}
\left\{\begin{matrix}
If & \frac{(z_x - \widetilde{x}_x(t+1))^2}{\sigma_{xx}^2}+\frac{z_y- \widetilde{x}_y(t+1) }{\sigma_{yy}^2} \leq 1 & \Rightarrow & w = \left |  \sqrt{\frac{(z_x - \widetilde{x}_x(t+1))^2}{\sigma_{xx}^2}+\frac{z_y- \widetilde{x}_y(t+1) }{\sigma_{yy}^2}} - 1   \right |   \\
If & \frac{(z_x - \widetilde{x}_x(t+1))^2}{\sigma_{xx}^2}+\frac{z_y- \widetilde{x}_y(t+1) }{\sigma_{yy}^2} > 1 &  \Rightarrow & w = 0
\end{matrix}\right.
\end{equation}
\begin{equation}
\label{eq:valorMedio}
z = \widetilde{x} + \frac{(z_1 - \widetilde{x}(t+1))w_1 + (z_2 - \widetilde{x}(t+1))w_2 +(z_3 - \widetilde{x}(t+1))w_3}{w_1 + w_2 + w_3}
\end{equation}

Fusing the set of available measurements provides the system with robustness and scalability. It is robust because, if~any of the developed measurements fail, the~system can continue to operate normally, and~it is scalable as other measurement systems are easy to integrate using the weighting philosophy described above. Furthermore, the~integrated measurement systems can be based on any of the available technologies, not only LiDAR. As~the number of measurements increases, the~result achieved should improve, considering the principles of Bayesian~statistics.

\section{Fail-Aware Odometry~System}
\label{sec:failAware}

The estimated time window evaluated by the fail-aware indicator is recalculated for each instant of time, allowing the trajectory planner system to manage an emergency manoeuvre in the best way. In~practice, most odometry systems do not implement this kind of indicator. Instead, our approach proposes the use of the evaluated heading error, as~the heading error magnitude is critical for the localisation error. Thus, a~small heading error at time $t$ produces a huge localisation error at time $t+N$ if the vehicle has moved hundreds of meters away. For~example, a~heading error equal to $10^{-3}$ rad at $t$ introduces a localisation error of 0.01 cm at $t+N$ if the vehicle moves only 100 m. This behaviour motivates us to use the heading error to develop the fail-aware~indicator.

The estimated heading error has a significant dependence on the localisation accuracy. The~developed fail-aware algorithm is composed of two parts: a function to evaluate the fail-aware indicator and a failure threshold, which is fixed as $0.001$. This threshold value was chosen by using heuristic rules and analysing the system behaviour in sequences 00 and 03 of the KITTI odometry data set. We evaluated the fail-aware indicator ($\eta$) on each odometry execution period, in~order to estimate the remaining time to overtake the fixed malfunction threshold. Equation~(\ref{eq:fail_aware}) defines the fail-aware indicator, where  $\sigma_{\psi\psi}$ is the estimated heading standard deviation and $\sigma_{\psi\psi}$ is identified as the variable most correlated with the localisation error; once again, regarding the error results in sequences 00 and~03.

For this reason, $\sigma_{\psi\psi}$ is useful to evaluate the fail-aware indicator. The~second derivative of $\sigma_{\psi\psi}$ is used, representing the heading error acceleration, so how fast or slow this magnitude changes is used as a determinant to find the estimated time of reaching the malfunction threshold. If~the acceleration of $\sigma_{\psi\psi}$ is low, the~estimated time window is large and the trajectory planner has more time to perform an emergency manoeuvre. On~the other hand, if~the acceleration of $\sigma_{\psi\psi}$  is high, the~estimated time window be decisive with respect to stopping the car safely in a short~time.

\begin{figure}[H]
\centering
\begin{subfigure}[]{.332\linewidth} \centering  \includegraphics[width = 0.9\linewidth] {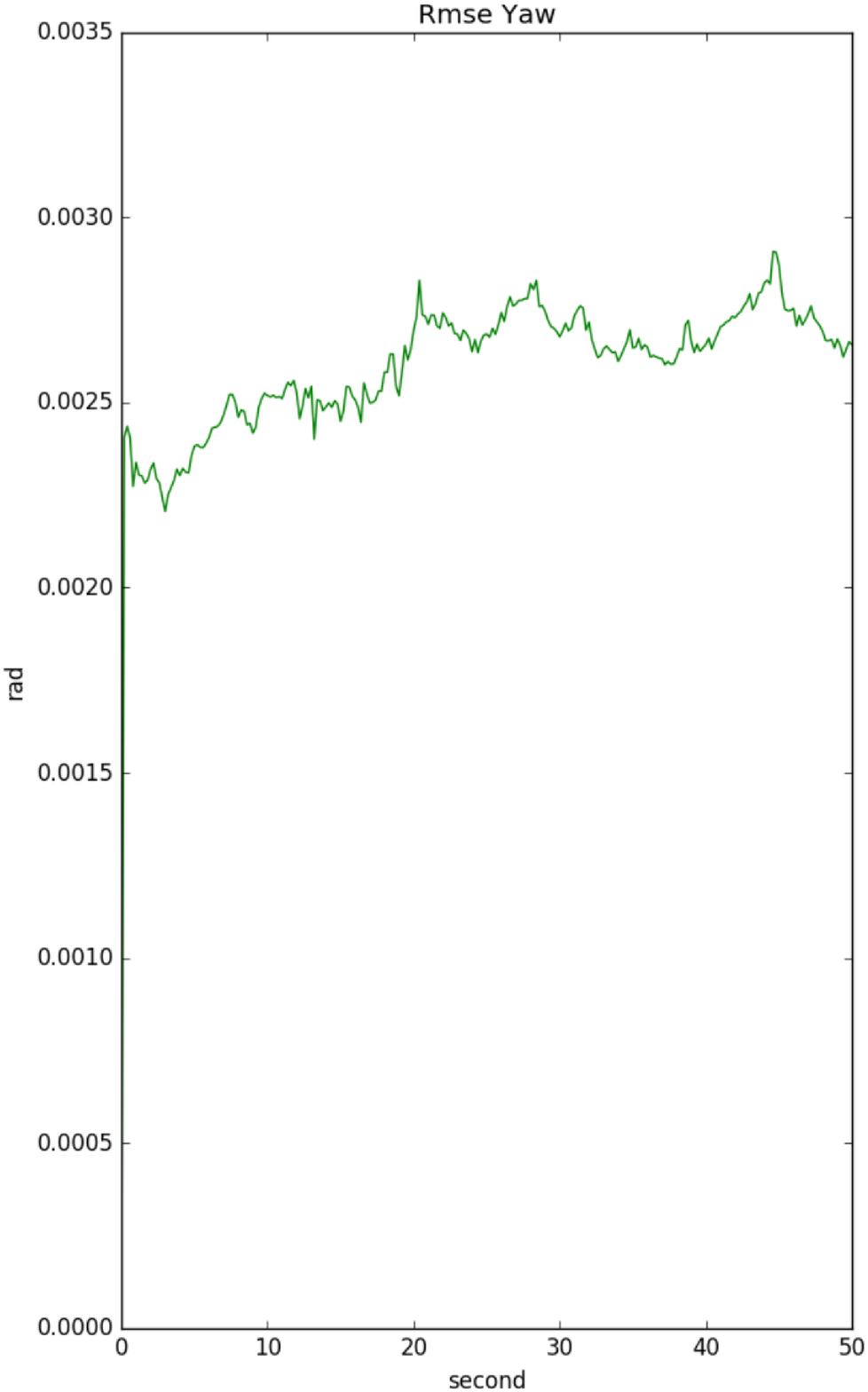} \caption{} \end{subfigure}
\begin{subfigure}[]{.33\linewidth} \centering  \includegraphics[width = 0.9\linewidth] {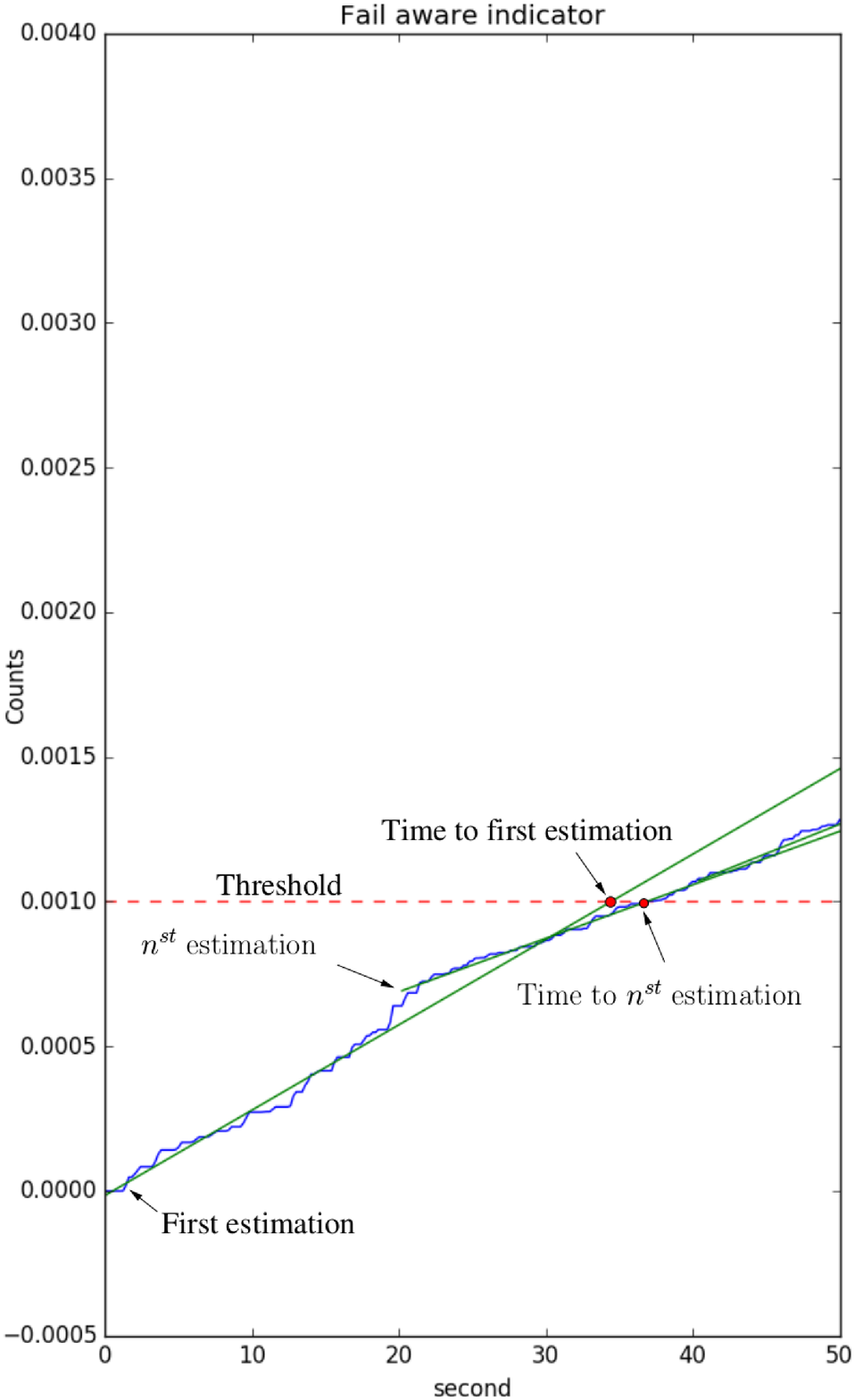} \caption{} \end{subfigure}
\begin{subfigure}[]{.32\linewidth} \centering  \includegraphics[width = 0.9\linewidth] {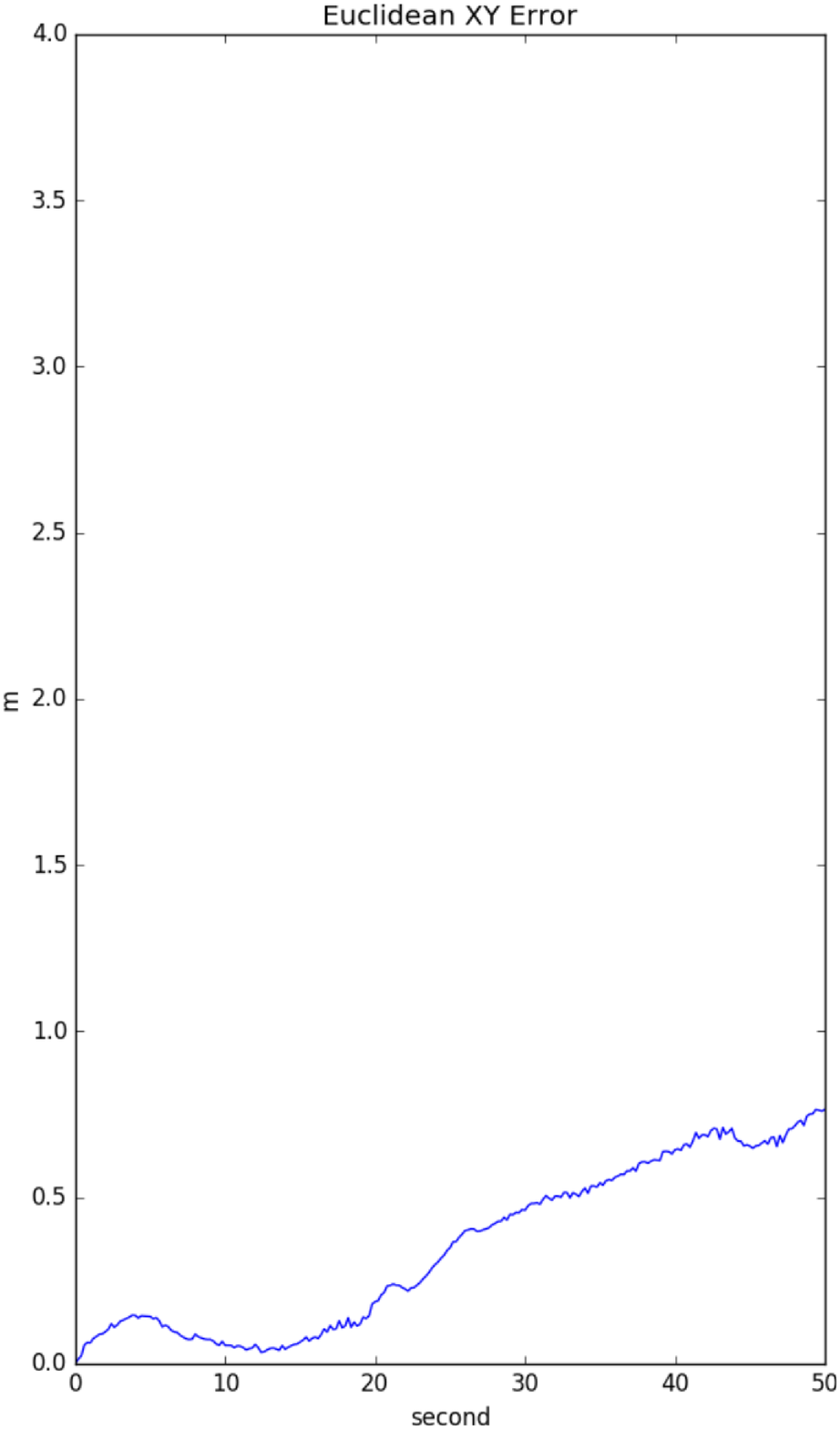} \caption{} \end{subfigure}
\caption{Fail-aware process. Sequence results 03. (\textbf{a}) Evolution of the signal standard deviation of $\widehat{\psi}$ estimated by the filter ($\sigma_{\psi\psi}$). (\textbf{b}) Representation of the failure threshold (red) and fail-aware indicator $\eta$ (blue). The~green lines represent the equation to evaluate the time window to reach the failure threshold. (\textbf{c}) Euclidean $[x, y]$ error compared with the ground truth (GT) of the data~set.}
\label{fig:failAwareIndicator}
\end{figure}

The acceleration of $\sigma_{\psi\psi}$ can be positive or negative, but~the main idea is to accumulate the absolute value for all the odometry interactions, in~order to have an indicator that allows us to know the estimated time window. The~limit time $t_1$ in the addition term represents when the LiDAR odometry system starts to work as a redundant system for localisation tasks. In~this way, the~speed $\eta$ is calculated as the difference between two consecutive $\eta$ values, in~order to assess the time to reach the malfunction threshold. Figure~\ref{fig:failAwareIndicator} shows the behaviour of the fail-aware algorithm. In~all the use-case studies, the~Euclidean error $[x, y]$ is approximated as 0.6 m when the malfunction threshold is exceeded. The~Euclidean XY error depicted in the image is calculated by comparing the LiDAR localisation and the available GT. The~fail-aware algorithm provides a continuous diagnostic value of the LiDAR system, allowing for the development of more robust and safe autonomous vehicles.
\begin{equation}
\label{eq:fail_aware}
\eta = \sum_{t=t1}^{\infty} \left \| \frac{d^2}{dt^2} \widehat{\sigma}_{\psi\psi}\right \|
\end{equation}

\unskip

\section{Experimental~Analysis}
\label{section:experimentalAnalysis}
\vspace{-6pt}

\subsection{KITTI Odometry Data Set~Evaluation}

\begin{figure}[H]
\centering
\includegraphics[width=0.95\linewidth]{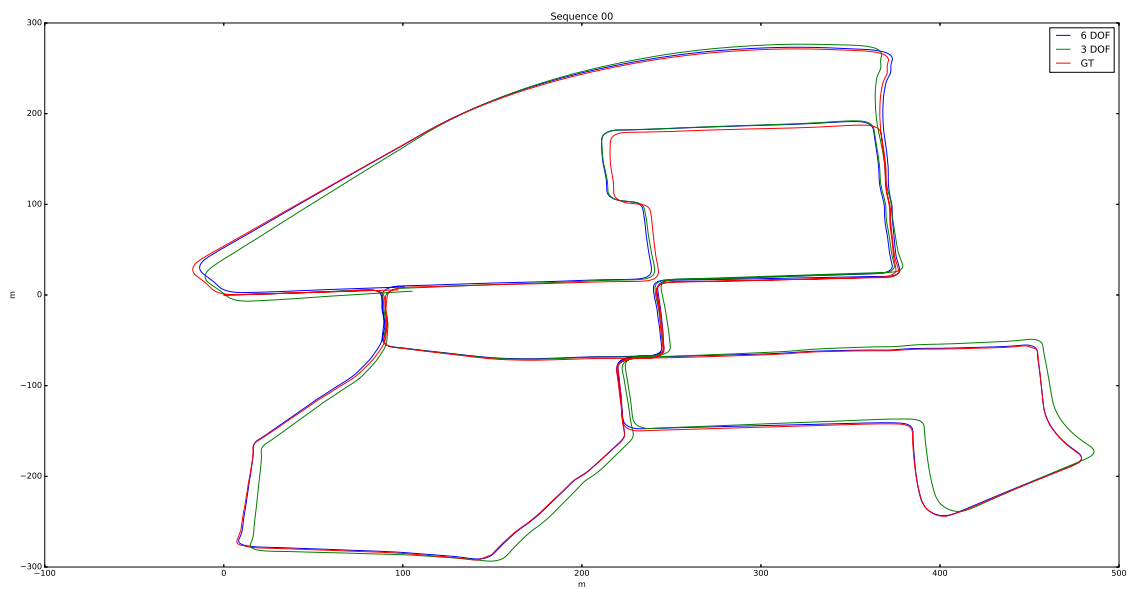}
\caption{Visual  results comparison using the 6-DoF and 3-DoF models in sequence~00.}
\label{fig:sequence_00_63DOF}
\end{figure}

The presented algorithm was extensively tested. A~total of approximately 50,000 point clouds from different environments and with a multitude of situations were processed, representing a total of 19.5 km processed. We defined four categories---urban, country, urban/country, and~highway---to~label each of the processed sequences. Table~\ref{tab:sequences} lists the translation and rotation errors obtained in each sequence as a result of applying~\cite{Geiger2012CVPR} for evaluation. Two use-cases, 6-DoF and 3-DoF, were evaluated to quantify the improvement introduced in the case of 6-DoF. The~results show that the odometry system worked for different scenarios, without~showing considerable differences in the results. However, sequence 01 (Highway) had considerable translation and rotation errors for the 6-DoF and 3-DoF cases, mainly because the road had a lower number of characteristics in the highway scenario. The~average results for 6-DoF were $1.00\%$ and 0.0039\: deg/m in translation and rotation, respectively. In~the case of 3-DoF, where the state vector $\widehat{x}(t)$ was defined by the variables $(x, y, \psi)$, the~mean error values were $7.79\%$ and 0.057 deg/m in translation and rotation, respectively. Figure~\ref{fig:sequence_00_63DOF} represents the results of processing sequence 00 in both~cases.

\begin{table}[H]
\caption{Numerical results when processing the sequences with 6-DoF or~3-DoF.}
\centering
\scalebox{0.95}[0.95]{\begin{tabular}{cccccc}
\toprule
\multirow{2}{*}{\textbf{Sequence}}  & \multirow{2}{*}{\textbf{Scene}} & \multicolumn{2}{c}{\textbf{6-DoF Error}}  & \multicolumn{2}{c}{\textbf{3-DoF Error}} \\\cmidrule{3-6}
&  & \textbf{Translational [\%]}  & \textbf{Angular [deg/m]} & \textbf{Translational [\%]}  & \textbf{Angular [deg/m]}\\
\midrule
00 & Urban       & 1.28 & 0.0051 & 9.87 & 0.0793  \\
01 & Highway         & 2.36 & 0.0135 & 12.89 & 0.0462 \\
02 & Urban/Country & 1.15 & 0.0028 & 4.42 & 0.0252  \\
03 & Country        & 0.93 & 0.0024 & 12.54 & 0.0864 \\
04 & Country        & 0.98 & 0.0033 & 1.34 & 0.0037  \\
05 & Urban       & 0.45 & 0.0018 & 10.01 & 0.0682 \\
07 & Urban       & 0.44 & 0.0034 & 3.39 & 0.0656  \\
09 & Urban/Country & 0.64 & 0.0013 & 3.84 & 0.0219  \\
10 & Urban/Country & 0.83 & 0.0017 & 12.29 & 0.0557 \\
\bottomrule
\end{tabular}}
\label{tab:sequences}
\end{table}
\unskip

On the other hand, the~results analysed were evaluated with the integration of three or two~measurements in the fusion system. The~three-measurement fusion combined the three techniques described in the article, while the two-measurement fusion only combined the ICP-based techniques. Table~\ref{tab:sequencesNoSVD} shows the translation and rotation errors for each sequence in both cases. In~the case of fusing only two measures, the~average result was $1.61 \%$ and 0.0046 deg/m in translation and rotation, respectively. Therefore, the~system with the three-measurement fusion improved the odometry behaviour by 62.3\%, as~compared to that with two-measurement~fusion.

\begin{table}[H]
\caption{Results of processed sequences with and without applying feature detection and SVD in the measurement fusion~process.}
\centering
\scalebox{0.95}[0.95]{\begin{tabular}{cccccc}
\toprule
\multirow{2}{*}{\textbf{Sequence}}  & \multirow{2}{*}{\textbf{Scene} }& \multicolumn{2}{c}{\textbf{Fusion with 3 Measures}}  & \multicolumn{2}{c}{\textbf{Fusion with 2 Measures}} \\\cmidrule{3-6}
&  & \textbf{Translational [\%]}  & \textbf{Angular [deg/m]} & \textbf{Translational [\%]}  & \textbf{Angular [deg/m]}\\
\midrule
00 & Urban         & 1.28 & 0.0051 & 1.31 & 0.0052 \\
01 & Highway       & 2.36 & 0.0135 & 7.08 & 0.0122 \\
02 & Urban/Country & 1.15 & 0.0028 & 1.21 & 0.0030 \\
03 & Country       & 0.93 & 0.0024 & 0.97 & 0.0022 \\
04 & Country       & 0.98 & 0.0033 & 0.69 & 0.0031 \\
05 & Urban         & 0.45 & 0.0018 & 0.91 & 0.0052 \\
07 & Urban         & 0.44 & 0.0034 & 0.63 & 0.0022 \\
09 & Urban/Country & 0.64 & 0.0013 & 0.93 & 0.0014 \\
10 & Urban/Country & 0.83 & 0.0017 & 0.84 & 0.0017 \\

\bottomrule
\end{tabular}}
\label{tab:sequencesNoSVD}
\end{table}

\begin{figure}[H]
\centering
\includegraphics[width=0.95\linewidth]{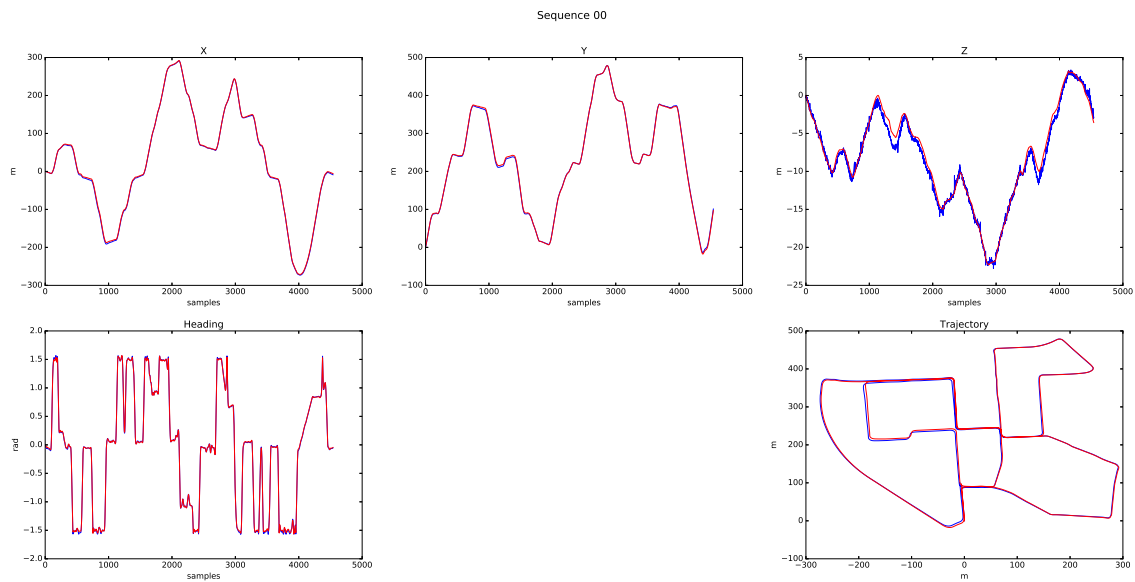}
\caption{{Sequence 00 results}.}
\label{fig:sequence00}
\end{figure}

One of the most well-known sequences in the KITTI odometry database is 00, as~it has been analysed and referenced in many SLAM and odometry papers, with~an approximate length of 3.8~km. Figure~\ref{fig:sequence00} shows the results of processing it, where the components $(x,y,z, \psi)$ of the estimated state vector $\widehat{x}$ are represented, as~well as the 2D path followed. All plots overlap the ground truth (GT) information with the odometry results. The~algorithm behaved properly visually, but~ended the sequence with error in all its variables: $error_x$ = 3 m, $error_y$ = 4 m, $error_z$ = 0.6 m, $error_{\alpha}$ = 0.02 rad, $error_{\theta}$ = 0.002 rad, and~$error_{\psi}$ = 0.007 rad.

\begin{figure}[H]
\centering
\includegraphics[width=0.95\linewidth]{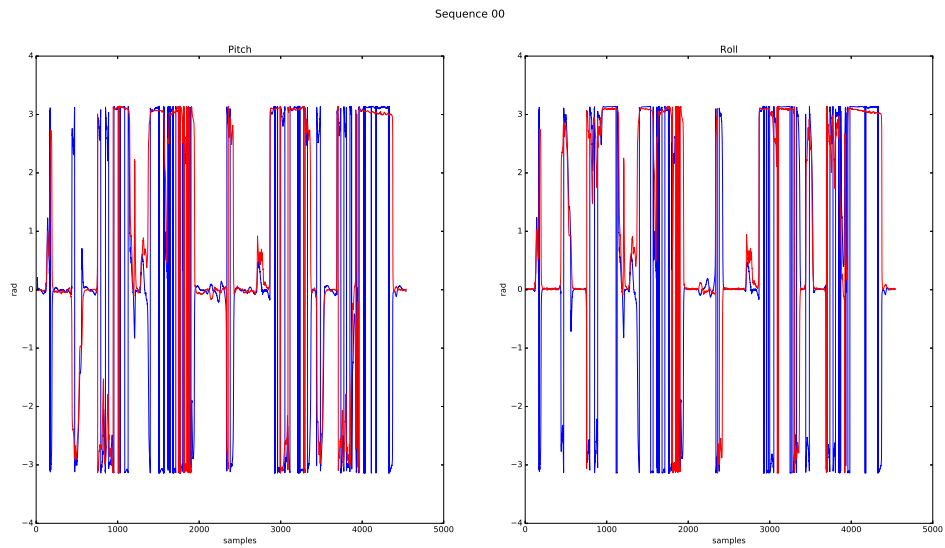}
\caption{Pitch and roll results for sequence~00.}
\label{fig:sequence00_pitchRoll}
\end{figure}

Figure~\ref{fig:sequence00_pitchRoll} shows the behaviour of the pitch and roll angles, where (concerning the previous representations) the similarities with the GT are not as evident, due to the angle normalisation done between $\pm \pi$, besides~putting in question whether the GT information was correct for the whole~sequence.

\begin{figure}[H]
\centering
\includegraphics[width=1\linewidth]{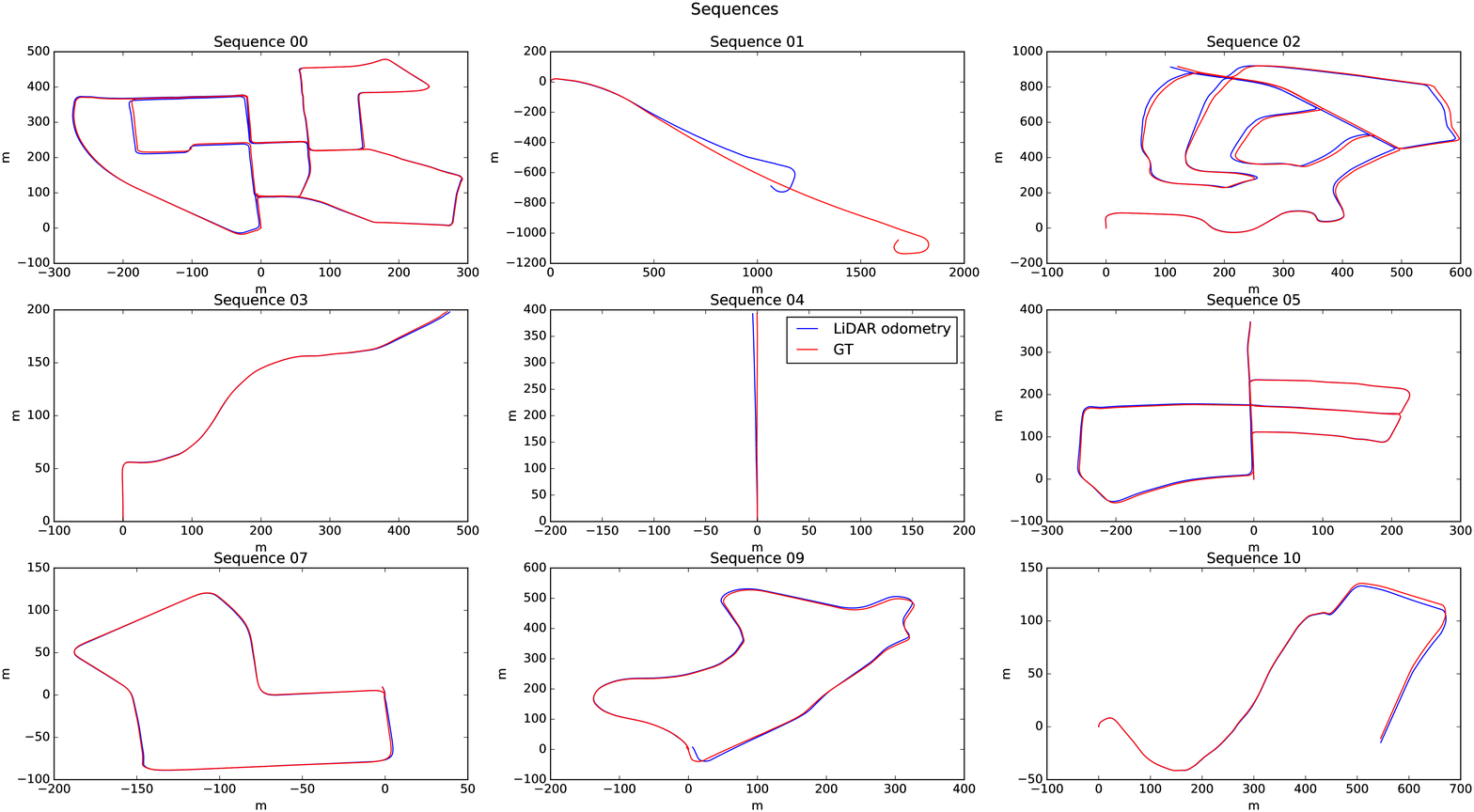}
\caption{Sequence~results.}
\label{fig:trajectorySequences}
\end{figure}

Figure~\ref{fig:trajectorySequences} shows the path of the sequences available in the database. Sequences 06 and 08 were removed from the results, as~the GT was not correct in both of these sequences. For~all cases, a~correct behaviour can be seen, except~for sequence 01, which shows significant error concerning the GT. This~error comes from the scenario in which the test was carried out, an~open road without objects, where characteristics could not be extracted and which lacked relevant points to apply the ICP techniques correctly. Errors were caused when a local minimum was detected, such that the integrated odometry process made the vehicle trajectory drift and increased the error in its~evaluation.

\begin{figure}[H]
\centering
\begin{subfigure}[]{.99\linewidth} \centering \includegraphics[width=1\linewidth]{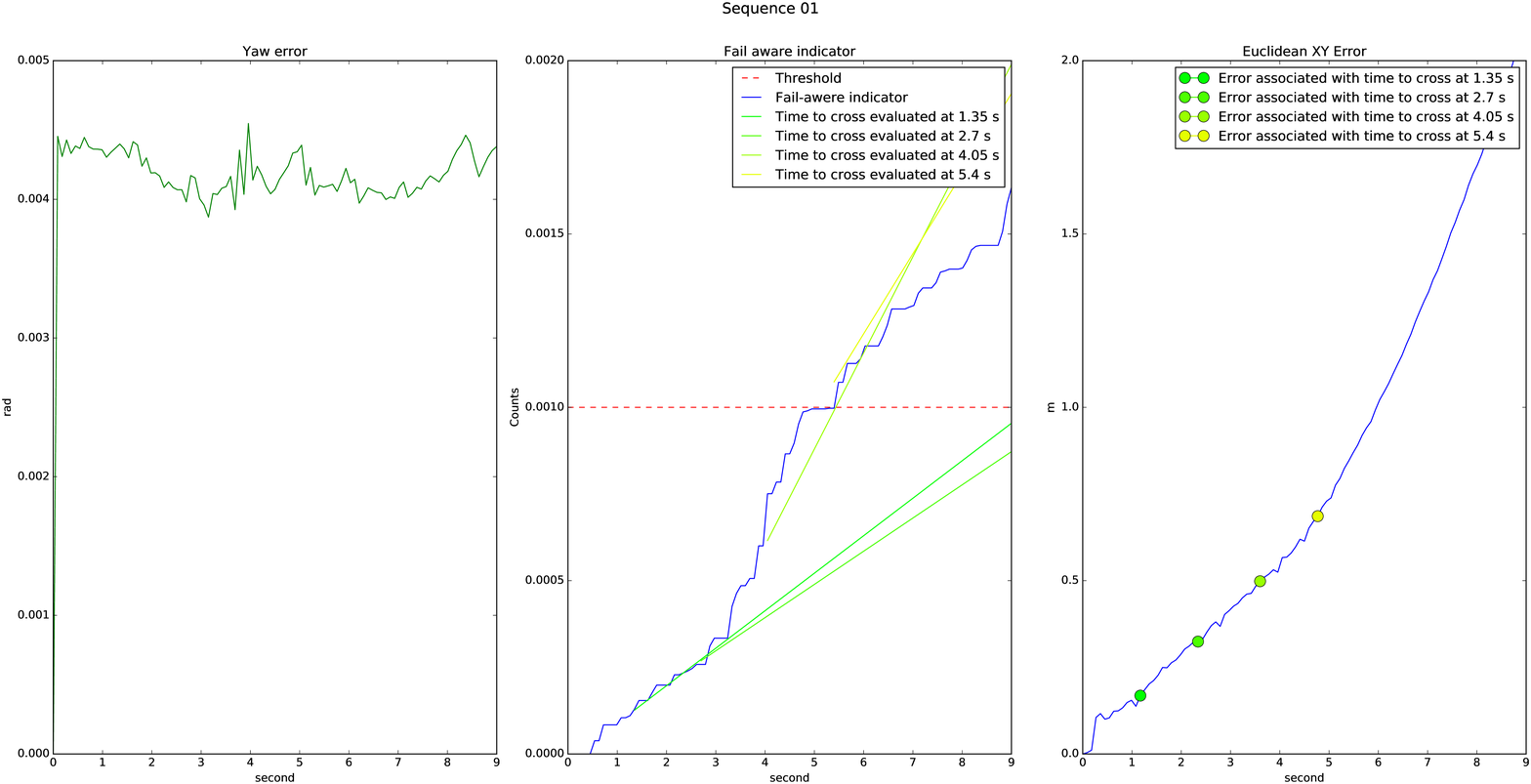} \caption{} \end{subfigure} \\\vspace{0.5cm}
\begin{subfigure}[]{.99\linewidth} \centering \includegraphics[width=1\linewidth]{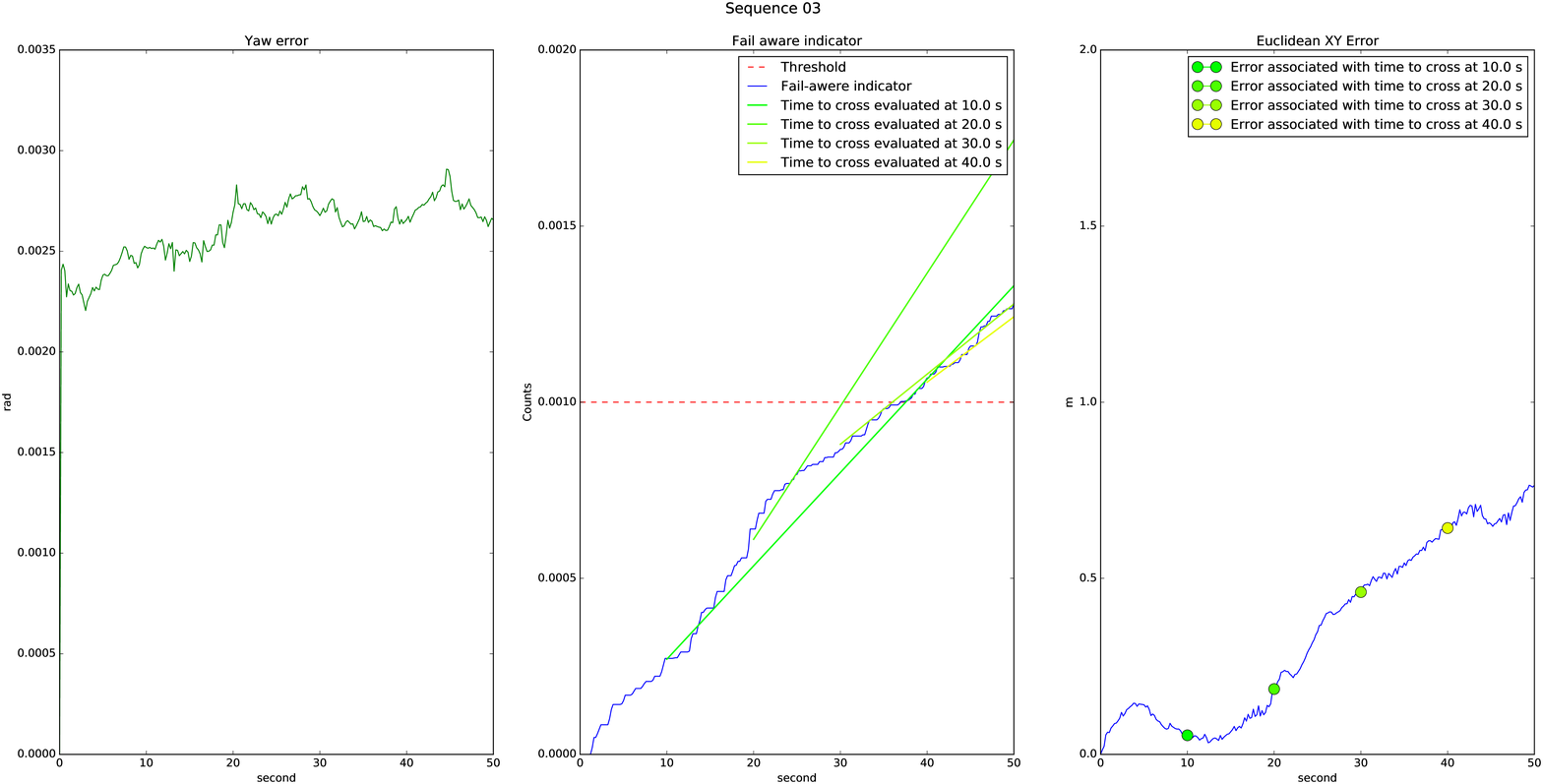} \caption{} \end{subfigure}
\caption{Dynamics of the fail-aware indicator: (\textbf{a}) Sequence 01 shows an estimated failure time \mbox{of 5~s}, which means that a high error is caused in a short time as a result of incorrect linear and angular localisation;  (\textbf{b}) Sequence 03 shows an estimated failure time of 40~s.}
\label{fig:failAware_sequence_01_04}
\end{figure}

The bad results shown in sequence 01 identify a system malfunction. The~fault was detected when processing the first point cloud, and~the fail-aware indicator showed a considerable difference from that in a sequence with acceptable results. Figure~\ref{fig:failAware_sequence_01_04}a shows the fail-aware indicator for sequence~01, as~compared with the indicator for sequence 03 shown in Figure~\ref{fig:failAware_sequence_01_04}b The analysis shows that the estimated time to reach the incorrect operation threshold was lower in sequence 01 than in sequence 03, with~values of 5 s and 40 s, respectively. Sequence 01 is characterised by a slow indicator change in its first seconds, estimating a failure time of approximately 10 s. However, after~three seconds of operation, the~indicator change increased considerably, reducing the failure time estimate to 5 s. The~times listed were taken with regards to the beginning of the test; although, in~a real scenario, these~times are relative to the current instant. An~estimated failure time of 5 s makes it practically impossible to carry out a safe stop manoeuvre. On~the other hand, the~speed of the indicator in sequence 03 is slow from the beginning of the test, and~continues with a similar speed until reaching the failure threshold. As~the indicator speed was slow, the~system could operate with an error of less than 0.5 m for approximately~40~s, allowing the planning system to carry out a safe stop~manoeuvre.

\begin{figure}[H]
\centering
\includegraphics[width=1\linewidth]{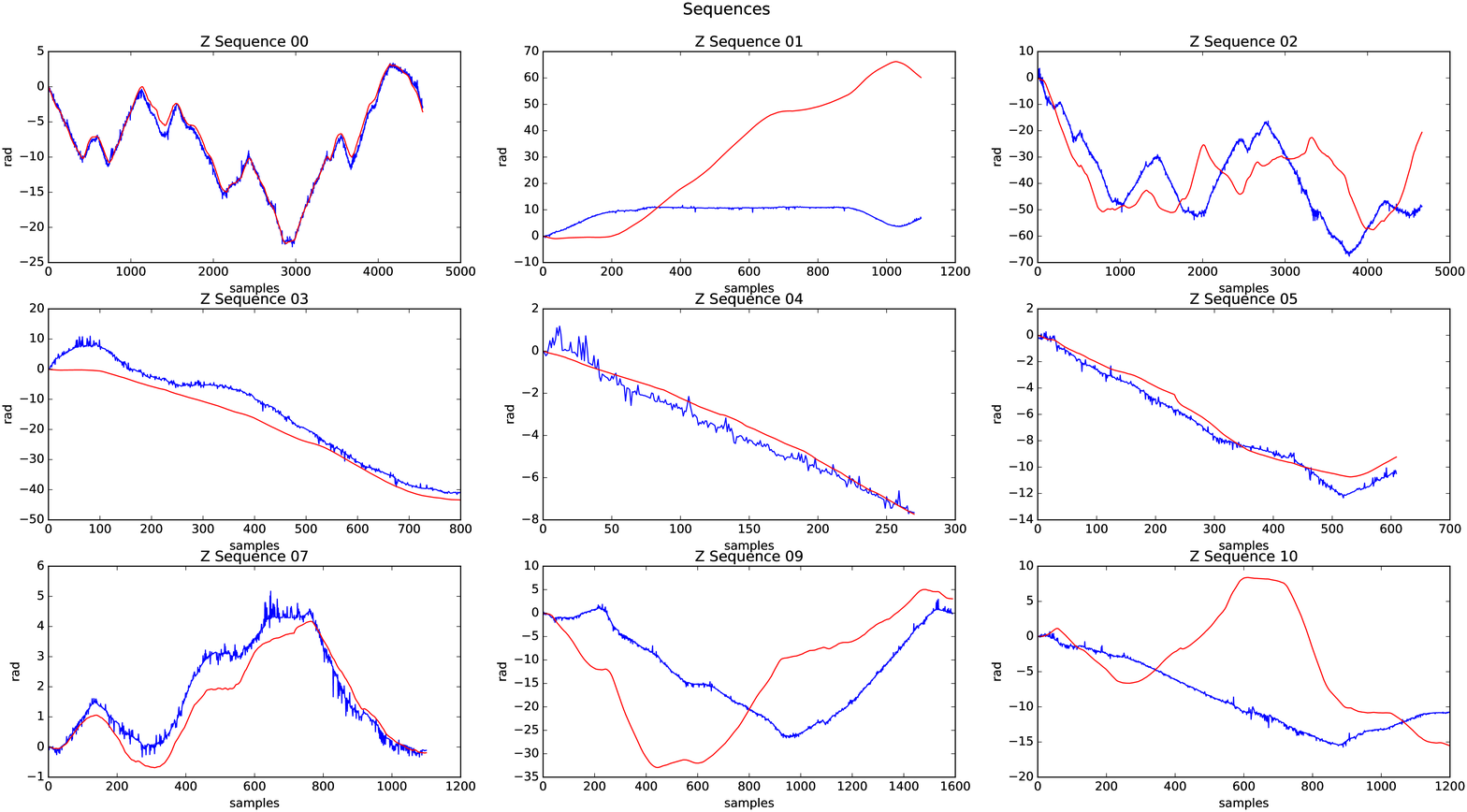}
\caption{Height sequences~results.}
\label{fig:zSequences}
\end{figure}

On the other hand, Figure~\ref{fig:zSequences} shows the estimated height of all processed sequences. The~good behaviour presented in the height estimation for six sequences should be noted. However, in~sequences 01, 09, and~10, the~estimated height was far from the GT. The~poor odometry behaviour justifies the error in 01, but~the errors in 09 and 10 are mainly due to the urban/country environment in which the sequences are developed, due to limited features to resolve the height~estimate.

\begin{figure}[H]
\centering
\includegraphics[width=1\linewidth]{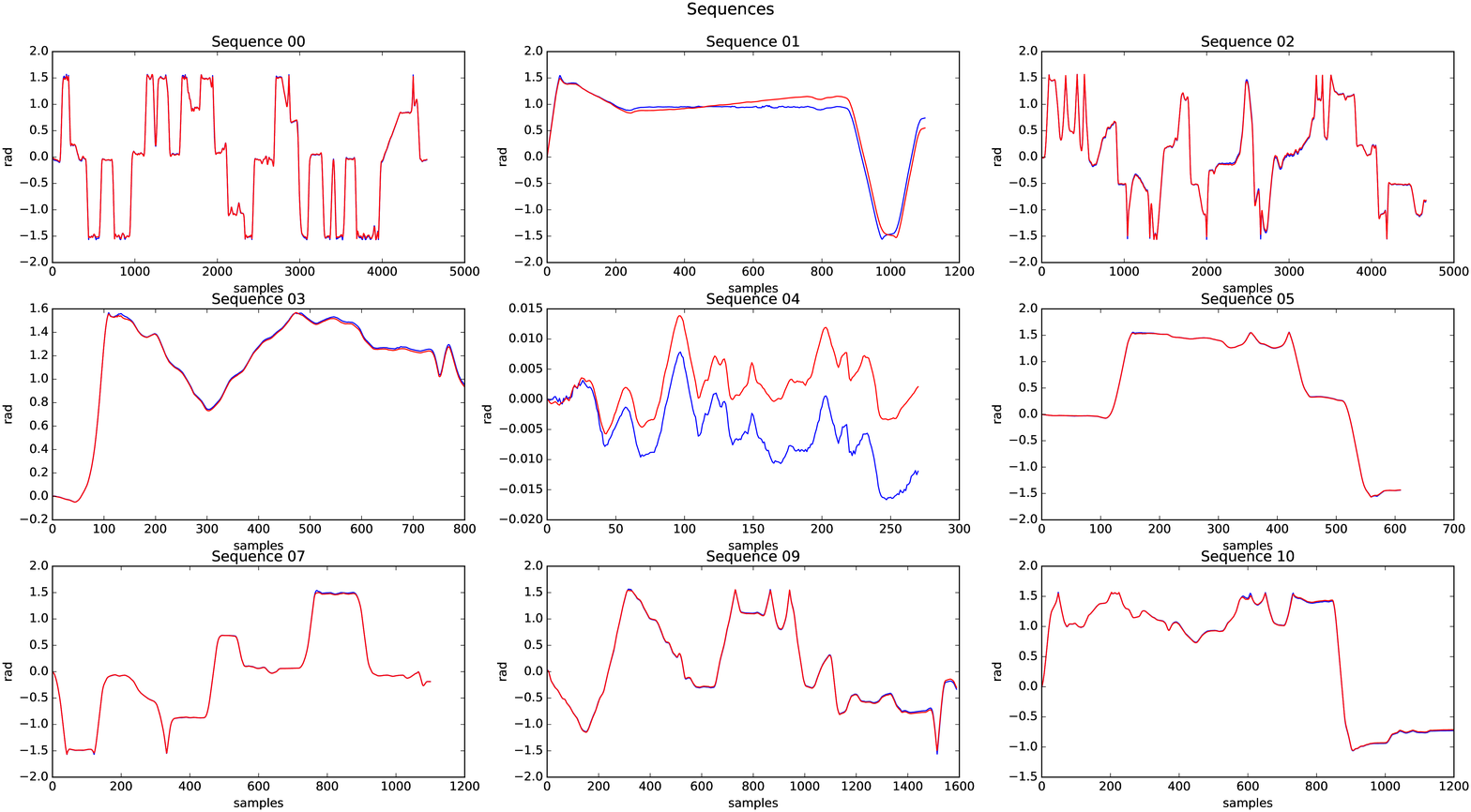}
\caption{Heading sequences~results.}
\label{fig:headingSequences}
\end{figure}

Finally, Figure~\ref{fig:headingSequences} shows the estimated heading of each of the processed sequences. It is essential to highlight the excellent behaviour presented by the algorithm in practically all of them. However, the~error at the end of 01 was 0.5 rad, much higher than that in the other sequences (i.e., close to 0.001~rad). This circumstance is essential to justify the excessive failure of~01.

\unskip

\subsection{Ranking~Evaluation}

In the set of listed algorithms in the KITTI odometry data set, there are a total of twenty-five pure LiDAR-based odometry algorithms, which we used to compare the results of the developed algorithm. Table~\ref{tab:ranking} shows the first fifteen entries of the ranking where, between~the LOAM and MDT-LO algorithms, the~translation error is doubled. The~implemented LiDAR odometry algorithm reached an average translation error of $1.00 \%$ and an average rotation error of 0.0039 deg/m; these~errors were obtained when processing the totality of sequences for which the GT was available. The~results achieved were encouraging, as~they were within the first fifteen entries in a data set that is well-known at an international level by the scientific community. The~ranked position was close to the 30th position, if~all entries in the data set are considered. We believe these are quite acceptable results, considering those systems that have the objective of being redundant localisation systems in autonomous driving. In~addition, the~fail-aware functionality of the system adds a differentiating feature from the state-of-the-art.
This feature allows the planning algorithms to compute safe trajectories even when the GPS localisation system fails,  simply by using only our odometry~method.

\begin{table}[H]
\caption{Comparison of LIDAR-based methods ranked on KITTI data~set.}
\centering
\scalebox{0.95}[0.95]{\begin{tabular}{ccccc}
\toprule
\textbf{Method}  & \textbf{KITTI Ranking Position} & \textbf{Translational Error [\%]}  & \textbf{Angular Error [deg/m]} & \textbf{Fail-Aware}\\
\midrule
LOAM~\cite{LOAM}       & 2 & 0.55 & 0.0013 & \\
IMLS-SLAM~\cite{imls}  & 4 & 0.69 & 0.0018 & \\
MC2SLAM~\cite{mc2}     & 5 & 0.69 & 0.0016 & \\
LIMO2-GP~\cite{limo}   & 11 & 0.84 & 0.0022 & \\
CAE-LO~\cite{caelo}    & 13 & 0.86 & 0.0025 & \\
LIMO2      & 15 & 0.86 & 0.0022 & \\
ICP-LO     & 16 & 0.87 & 0.0036 & \\
CPFG~\cite{cpfg}  & 17 & 0.87 & 0.0025 & \\
PNDT LO~\cite{pndt}   & 22 & 0.89 & 0.0030 & \\
LIMO      & 24 & 0.93 & 0.0026 & \\
S4-SLAM    & 28 & 0.98 & 0.0044 & \\
RIS        & 29 & 0.98 & 0.0026 & \\
\textbf{Ours} & - & 1.00 & 0.0039 & x \\
KF-SLAM    & 30 & 1.00 & 0.0041 & \\
S4OM       & 34 & 1.03 & 0.0053 & \\
NDT-LO     & 35 & 1.05 & 0.0043 & \\
\bottomrule
\end{tabular}}
\label{tab:ranking}
\end{table}
\unskip

\section{Conclusions and Future~Works}
\label{sec:conclusions}

Although research and development into autonomous driving in recent decades has helped to achieve high SAE levels of automation, the~presently existing control architectures are highly driver-dependent. Indeed, in~the case of hardware failure, the~driver must take control of the vehicle. However, to~improve user acceptance of these systems, the~vehicle itself should be able to solve the problem autonomously.
This paper presents a LiDAR odometry system with an integrated fail-aware feature, which notifies high-level systems with the actual performance of our proposal.
For instance, this allows a trajectory planner to plan a safe stop manoeuvre, guaranteeing the needed security in the~environment.

In this paper, we presented a robust and scalable localisation system which, independently of the support or redundancy that it can offer to other systems, allows  for fusion with any of the available localisation measures, improving its robustness in the measure. Thus, the~fusion architecture presented reduces the undesirable consequences given in urban scenarios where the D-GPS measure may suffer from loss of accuracy due to satellite~visibility.

Our proposal is based on a LiDAR-based odometry algorithm. An~in-depth study of the topic was presented and it was pointed out that all existing odometry systems suffer from drift error in their process.
To minimize and possibly eliminate this issue, the~proposed system fuses different LiDAR-based localisation measurements using a UKF filter. The~filter implementation takes into account 6-DoF dynamic models, which improve the correction process of the point cloud sweep, correcting the deformation of the scene when it is captured from a moving platform with high angular velocities. The~model also allows us to estimate the vehicle roll and pitch variables, in~order to reduce the measurement~noise.

ICP techniques have been widely applied in LiDAR odometry systems, but~these techniques have the disadvantage of being slow, despite their high precision. Therefore, the~presented approach integrates multiple measurement stages to improve the accuracy of the final measurement. The~first stage is based on multiple ICPs of lesser precision but with starting seeds distributed to improve its precision in the final measurement. The~second stage is based on applying constraints to the ICP minimisation process, improving the accuracy and the computation time. Finally, the~third stage is focused on the detection of vertical corner features above the point clouds, in~order to apply SVD and estimate the homogenous transform between point clouds. However, the~results of the third stage are conditioned to the type of scene, as~shown in the highway results of Sequence~01.

The obtained linear and angular errors when processing the KITTI odometry data set were $1.00 \%$ and 0.0039 deg/m, respectively. These results are ranked within the first fifteen methods based only on LiDAR odometry. Furthermore, the~proposed algorithm introduces a dynamic fail-aware indicator, a~function of the standard error deviation associated with the estimation of the yaw vehicle~angle.

As this work presents a fail-aware system based on LiDAR odometry, it could assist other systems of the vehicle (which is part of our planned future work) to decrease the linear and angular errors associated with localisation. For~this purpose, a~new localisation measure based on semantic segmentation and machine learning techniques should be added. On~the other hand, building high-definition 3D maps is a booming topic, which may solve many of the problems that autonomous driving is prone to at present. Therefore, integrating a global D-GPS measurement into the developed system may eliminate the drift, allowing us to build high-definition 3D~maps.


\vspace{6pt}




\authorcontributions{Conceptualization, I.G.D. and M.R.; methodology, I.G.D.; software, I.G.D. and M.R.; validation, M.R., R.I.G., and~C.S.M.; formal analysis, I.G.D.; investigation, I.G.D. and M.R.; resources, N.H.P.;  writing---original draft preparation, I.G.D.; writing---review and editing, C.S.M., A.B., and~N.H.P.; visualization,~N.H.P.; supervision, D.F.L. All~authors have read and agreed to the published version of the manuscript.}

\funding{This work was supported in part by grant CCG2018/EXP-065 (UAH), in part by grant S2018/EMT-4362 SEGVAUTO 4.0 (CAM), in part by grant DPI2017-90035-R (Spanish Ministry of Science and Innovation), in part by grant 723021 (BRAVE, H2020), in part by the Electronic  Component Systems for European Leadership Joint Undertaking through the European Union’s Horizon 2020 Research and Innovation Program and Germany, Austria, Spain, Italy, Latvia,  Belgium, The Netherlands, Sweden, Finland, Lithuania, Czech Republic, Romania, and Norway, under Grant 737469, in part by grant TRA2017-90620-REDT (Spanish Ministry of Science and Innovation) and in part by grant agreement Marie Sk\l{}odowska-Curie 754382.}


\conflictsofinterest{The authors declare no conflict of~interest.}
\abbreviations{The following abbreviations are used in this manuscript:\\

\noindent
\begin{tabular}{@{}ll}
GPS & Global Position System\\
DoF & Degrees Of Freedom\\
ICP & Iterative Closest Point\\
IMU & Inertial Measurement Unit\\
KITTI & Karsruhe Institute of Technology and Toyota Technological Institute\\
LiDAR & Laser imaging Detection And Ranging\\
LOAM & LiDAR Odometry And Mapping in real time\\
RANSAC & RANdom Sample Consensus\\
SLAM & Simultaneous Localisation And Mapping\\
SVD & Singular Value Decomposition\\
UKF & Unscented Kalman Filter\\
\end{tabular}}

\appendixtitles{no} 


\reftitle{References}


\externalbibliography{yes}

\end{document}